\documentclass{article} % For LaTeX2e
\usepackage{vxw_style}
\usepackage{vxw_macros}
\usepackage{macros}
\arxivtrue

\let\citet\cite

\title{Mirror Descent at the Edge of Stability}
\author{
  Vincent X. Wang \\
  University of Chicago \\
  \texttt{vincentxwang@uchicago.edu}
  \and
  Vikas Garg \\
  Aalto University and YaiYai Ltd. \\
  \texttt{vgarg@csail.mit.edu} \\
}

\begin{document}

\maketitle

\begin{abstract}
In many deep learning settings, gradient descent has been shown to operate at the \emph{edge of stability} -- a regime where the largest eigenvalue of the loss Hessian hovers around $2/\eta$. We study this phenomenon in \emph{mirror descent}, a generalization of gradient descent. As mirror descent is non-Euclidean, existing explanations of the edge of stability phase do not extend to this setting. We generalize the existing notion of sharpness to introduce \emph{mirror sharpness}, which measures the curvature of the loss with respect to the mirror potential. Empirically, we observe that the mirror sharpness hovers slightly above the $2/\eta$ threshold. Furthermore, we derive a dual central flow ODE for mirror descent that models iterates in continuous time at the onset of edge of stability. Our ODE provides insights into the difficulties of extending existing arguments for why sharpness stabilizes at $2/\eta$ from gradient descent to mirror descent in full generality.

\end{abstract}

\section{Introduction}

First-order methods, such as gradient descent, are the de facto optimization methods for much of deep learning. Despite their empirical successes, their behaviors on realistic \emph{non-convex} objectives, such as neural network losses, are poorly understood and significantly depart from the classical picture of convex optimization.

For instance, the famed descent lemma \citep{Nes04} guarantees that if the maximum Hessian eigenvalue is uniformly bounded above by $M$, then gradient descent monotonically decreases the loss when the step size satisfies $\eta < 2/M$. However, in realistic settings, gradient descent has been shown to operate close to the boundary of this ``stable regime.'' In particular, \citet{CKLKT21} observed across a variety of architectures that, for a fixed step size $\eta$, the \emph{sharpness}, $S(x):=\maxEV{\nabla^2 L(x)},$ progressively increases until it reaches approximately $2/\eta$, after which it oscillates near this threshold while optimization continues. The first phase is known as \emph{progressive sharpening}, and the second is termed the \emph{edge of stability} (EoS).

While full-batch gradient descent is typically the testbed for the analysis of the EoS regime, recent works have shown that EoS also manifests in several fundamental optimizers beyond this setting. A key thread through some of these works is that the relevant notion of sharpness is connected to the geometry induced by the optimizer. For instance, \citet{CGKAM+23} formulates a \emph{preconditioned sharpness} and observes an EoS phase in adaptive preconditioning methods like AdaGrad and Adam. Most recently, \citet{ICCG26} formulates a \emph{generalized sharpness} and observes EoS behavior in steepest descent methods in arbitrary norm.

In this paper, we consider EoS behavior in mirror descent, a non-Euclidean extension of gradient descent that fundamentally differs from the steepest descent studied in \citet{ICCG26}. Whereas steepest descent ``fixes'' a geometry, mirror descent introduces a Bregman geometry that can change at every point. Furthermore, mirror descent differs from general adaptive preconditioning optimizers (such as AdaGrad and Adam, as studied in \citet{CGKAM+23}) because the Bregman geometry is \emph{explicitly} given by the iterate and does not otherwise depend on the history of the trajectory.

We believe the study of mirror descent to be important for two main reasons. First, we view understanding this regime to be a first step in building towards a theory that explains the \emph{mechanisms} of EoS in adaptively preconditioned optimizers (such as those in \citet{CGKAM+23}). While these optimizers differ from mirror descent, historically, the developments of several adaptive preconditioning methods like AdaGrad \citep{DHS11} and Shampoo \citep{GKS18} are based on the lens of (online) mirror descent. Second, mirror descent is a relatively tractable testbed for understanding fundamental differences in the EoS phenomenon between Euclidean and non-Euclidean regimes, as the Bregman geometry is well-structured and the update is entirely explicit.

In (stochastic) gradient descent, why does sharpness tend to hover around $2/\eta$? Several existing explanations of this phenomenon for gradient descent \citep{ALP22, DNL23, CDTKL24, LKK26} rely on the observation that at EoS, the gradient descent iterates oscillate along ``unstable'' eigendirections $u_k$ with eigenvalue $2/\eta$, and these oscillations induce a ``self-stabilizing'' effect. This understanding of optimization is formalized in the central flows framework of \citet{CDTKL24}, but it is unclear the extent to which this model holds for mirror descent. We find that such an explanation is \emph{not} sufficient to justify the self-stabilization phenomenon in mirror descent in full generality, and we believe that further exploration would clarify the extent to which self-stabilization is a universal phenomenon across non-Euclidean optimizers.

Our contributions are as follows:
\begin{itemize}
    \item In \Cref{sec:sharpness}, we define \emph{mirror sharpness}, which generalizes existing notions of sharpness to the point-wise geometry induced by mirror descent. We connect this definition to an approximate descent lemma (\Cref{prop:descent}) as well as optimization over a quadratic (\Cref{prop:quadratic_sharpness}).
    \item In \Cref{sec:mirror_stabilizes}, across a series of mirror maps and architectures, we empirically observe that mirror sharpness hovers around $2/\eta$. 
    \item In \Cref{app:mirror_flow}, we derive a central flow ODE that empirically describes the time-averaged iterates at the onset of EoS via the method of \citet{CDTKL24}, improving over Riemannian gradient flow. 
\end{itemize}

\subsection{Related Works}

\paragraph{Edge of stability} 
Even in the simplest case of full-batch gradient descent, the EoS phenomenon remains poorly understood. Many works attempt to study EoS dynamics in the case of specific simplified loss functions \citep{ZWWZG22, ABCLS+23, KNSC23, GKWRQ24, LZDZ25, YSY25}, even those beyond neural network losses. Another line of work \citep{ALP22, DNL23, CDTKL24, MMPTX+25, LKK26} analyzes EoS behavior for general losses. To our knowledge, all works of this sort exploit the \emph{Euclidean} fact that the gradient of the sharpness can be expressed with the third derivative of the loss, which \citet[Lemma 2]{DNL23} dubs the ``self-stabilization property.'' Unfortunately, for mirror sharpness, such a simple identity does not hold, and we find an additional third-order geometry correction (see \Cref{lem:sharp_grad}). This creates difficulties in analysis, which we make explicit in \Cref{sec:sometimes_OK}. 

\paragraph{Central flows}
Central flows, introduced by \cite{CDTKL24}, are ODEs that model optimization algorithms in continuous time. They differ from ``small step size'' approximations (e.g., Riemannian gradient flow) because they use the EoS phase as an ansatz to construct a flow. Some recent work propose modifications to the central flow \citep{RC26A, Mar26} and \citet{RC26B} extends a model to algorithms with momentum. We extend the central flow model of \cite{CDTKL24} to mirror descent in the regime where one eigenvalue is at $2/\eta$ by modeling oscillations in the \emph{dual} space, which we empirically show to track the mirror descent iterates better.

\paragraph{Implicit bias of mirror descent}
Mirror descent, introduced in \citet{NY83}, has been a central first-order optimization paradigm for decades, as it is a surprisingly broad framework. Recent works within the past decade have focused on the \emph{implicit bias} of mirror descent. For instance, in various regression settings, \citet{GLSS18, AH18, ALH22} argue that mirror descent finds a solution close to initialization in Bregman divergence. \citet{SATA22, SGAA23} show that in classification settings, mirror descent (with an $\ell_p$ potential) finds a solution that maximizes a margin with respect to an $\ell_p$ norm. Recent work \citep{PDF24, JM26} has also studied the implicit bias of the associated flow with mirror descent in various settings. Our paper instead studies the implicit bias of mirror descent through the complementary lens of sharpness; we think of the EoS phenomenon as implicitly induced through the discrete nature of mirror descent.

\paragraph{Applications of mirror descent}
In practice, mirror descent is most often used in optimization settings with constraints, although we do not deal with this topic in our paper. For instance, many policy optimization problems in reinforcement learning (e.g., LLM post-training) deal with optimizing over a simplex; thus, it is natural to approach these in terms of mirror descent. This view is made explicit in the framework of mirror descent policy optimization (MDPO) \citep{TSEG21}. Furthermore, popular algorithms such as TRPO \citep{SLAJM15} can also be interpreted as an approximate mirror descent step, as shown by \citet{NJG17} and \citet{GSP19}. Mirror descent variants are also frequently used in the min-max problem setting, which may arise when optimizing data mixtures for pretraining \citep{XPDDL+23} and aligning language models to human preferences \citep{munosNashLearningHuman2024, FAG26}.

\section{Background}
\label{app:bg}

\ifarxiv\else
We use standard notation, detailed in \Cref{sec:annoying_tensors}.
\fi
We will work with multi-linear operators though they are not required to understand the high-level contents of the paper. Throughout this paper, unless otherwise indicated, we will use $H(x) = \nabla^2L(x)$ and $G(x) = \nabla^2 \Phi(x)$, where $L$ is the loss and $\Phi$ is the mirror potential (as defined below). \textbf{In all formal statements and definitions, we will assume $L, \Phi$ to have appropriate regularity where needed.} Assuming that these are $C^3$ is usually sufficient (unless otherwise noted). We study the task of minimizing $L$ given queries to $\nabla L$.

% arXiv only: notation lives here; in the ICLR version it is Appendix A (multilinear_algebra.tex).
\ifarxiv
\paragraph{Notation} To make the distinction between primal coordinates and dual coordinates more clear, we will work over the parameter space $E = \mathbb{R}^n$ with dual $E^*$. Let $L: E \to \mathbb{R}$ be a $C^3$ loss function throughout this paper. We use $A \succ 0$ to denote that $A$ is positive definite. For a matrix $A$, let $\maxEV{A}$ to denote the largest eigenvalue of $A$, and let $\minEV{A}$ denote the smallest. We use $\|x\|_2^2 = \sum x_i^2$ to denote the squared $\ell_2$ norm.

\paragraph{Derivatives} In deriving the central flows for mirror descent, we will often use higher-order derivatives that are multilinear. We try to keep such notation at a minimum, opting to use the more familiar gradient $\nabla^k$ where possible. We refer the reader to \citep[Chapter XIII, \S6]{LangRFA} for explicit constructions of these objects.

Let $F: E \to E$ be a $C^k$ function (such as the loss $L$ or mirror potential $\Phi$). We use $D^kF(x): E^k \to E$ to denote the derivative viewed as a $k$-linear map, which is symmetric since we assume $F$ is $C^k$. We will frequently invoke Taylor's theorem; a proof can be found in \citep[Chapter XIII, \S6]{LangRFA}.

\begin{theorem}[Taylor's theorem]
    Let $f: E \to E$ be $C^p$. Denote by $y^{(k)}$ the $k$-tuple $[y,\dots, y]$. Then
    \[f(x + y) = f(x) + \sum_{k=1}^{p-1} \frac{D^{k}f(x)[y^{(k)}]}{k!} + O(\|y\|^p).\]
\end{theorem}

In the case when $f: E \to \mathbb{R}$ (as in the case of the loss $L$ or mirror potential $\Phi$) and $f$ is $C^k$, we opt to use the more familiar gradient notation $\nabla^k$. We will only invoke the $k\leq3$ cases. As the $k=1$ case corresponds to the gradient and the $k=2$ case corresponds to the Hessian, we will describe the $k=3$ case. Let $\nabla^3 f(x): E^2 \to E^*$ be the symmetric, bilinear operator that satisfies
\[\langle\nabla^3 f(x)[u,v], w \rangle = D^3 f(x)[u,v,w]\]
for all $u,v,w \in E$.

\fi

\paragraph{Generalized eigenvalues} We say $\lambda_i$ is a \emph{generalized eigenvalue}\footnote{This should not be confused with the definition of a generalized eigenvector as a vector $w$ such that $(A-\lambda I)^kw = 0$ for some $k \geq 1$ (i.e., those eigenvectors that form the Jordan decomposition).} of $(H, G)$ if $Hv_i = \lambda_iGv_i$ for some nonzero eigenvector $v_i$. When $G$ is the identity, this reduces to an eigenvalue of $H$.

\paragraph{Mirror descent} We say a function $\Phi: E \to \mathbb{R}$ is a \emph{mirror potential} if it is differentiable, strictly convex, and its gradient $\nabla \Phi: E \to E^*$ is a bijection. We call $\nabla \Phi: E \to E^*$ the \emph{mirror map}. Its inverse is given by $(\nabla \Phi)^{-1} = \nabla \Phi^*$, where $\Phi^*$ denotes the convex conjugate of $\Phi$. In this paper, we set $E = \mathbb{R}^n$.\footnote{This is not the case for certain choices of mirror descent (most notably, the multiplicative weights update induced by the negative entropy $\Phi_{\text{ent}}(x) = \sum_{i} x_i \log x_i$).}

The natural notion of ``distance'' induced by $\Phi$ on $E$ is the \emph{Bregman divergence} $D_{\Phi}(x,y) := \Phi(x) - \Phi(y) - \langle \nabla \Phi(y), x - y \rangle$. Note that the Bregman divergence is not symmetric (i.e., in general, $D_\Phi(x,y) \neq D_\Phi(y,x)$). When $\Phi$ is strongly convex, $G$ is a positive-definite metric (i.e., $G(x) \succ 0$ for all $x \in E$).

The mirror descent (MD) algorithm is given by the update
\[x_{t+1} = \argmin_{y \in E } \{  \eta \langle \nabla L(x_t), y - x_t \rangle + D_{\Phi}(y, x_{t}) \},\]
or equivalently
\begin{equation}
\label{eq:MD}
    x_{t+1} = \nabla\Phi^*(\nabla \Phi(x_t) - \eta \nabla L(x_t)). \tag{MD}
\end{equation}
Conceptually, we may also think of the above as updates in dual coordinates $z_t = \nabla \Phi(x_t)$ given by
\[z_{t+1} = z_{t} - \eta (\nabla L \circ \nabla \Phi^*)(z_t).\]
We may refer to both $z_t$ and $x_t$ as iterates because the mirror map $\nabla \Phi$ is bijective. As an example, when $\Phi(x) = \frac{1}{2} \|x\|_2^2$, we retrieve ordinary gradient descent, with $x_t = z_t$.

\section{Defining sharpness}
\label{sec:sharpness}

In this section, we generalize the notion of sharpness to mirror descent. Recall that for ordinary gradient descent, the sharpness is given by 
\[S^{\ell_2}(x) := \maxEV{\nabla^2 L(x) } = \max_{\|u\|_2 = 1} \langle u, \nabla^2 L(x) u \rangle.\]
A loose geometric interpretation of this quantity is the maximum curvature\footnote{More precisely, we use curvature to mean how curved the quadratic upper bounding $L$ must be.} of $L$ along any direction $u$. We will refer to this notion of sharpness as the $\ell_2$ sharpness.

Our definition is inspired by the work of \cite{GWS21}, which interprets mirror descent \emph{in the primal space} as a partial discretization of a Riemannian gradient flow with geometry $G(x) := \nabla^2 \Phi(x)$. Thus, it is natural to measure the curvature of $L$ under this geometry:

\begin{definition}[Mirror sharpness]
\label[definition]{def:sharpness}
Let $\nabla^2 \Phi (x) \succ 0$. For $x \in E$, the \emph{mirror sharpness}\footnote{More generally, one may define the sharpness with respect to an arbitrary metric $G$ by replacing $\nabla^2 \Phi(x)$ with $G(x)$.} $S^\Phi: E \to \mathbb{R}$ is given by
\begin{align*}
    S^{\Phi}(x) &:= \max_{\langle u, \nabla^2 \Phi(x)u \rangle = 1} \langle u, \nabla^2 L(x) u\rangle. 
\end{align*}
Equivalently, $S^\Phi(x)$ is the top generalized eigenvalue of $(\nabla^2 L(x), \nabla^2 \Phi(x)).$
\end{definition}

Similar to the definition for $\ell_2$ sharpness, $S^\Phi$ is the maximum curvature of $L$ along any direction $u$ \emph{with respect to the local metric $G(x) = \nabla^2 \Phi(x)$}.\footnote{Our notion of mirror sharpness is closely related to work of \citet{LFN18}, which connects a similar Hessian formulation to \emph{relative smoothness} coefficients. An important difference is that our definition applies to a point locally rather than globally.} Note that when $\Phi(x) = \frac{1}{2}\|x\|_2^2$, we retrieve $\ell_2$ sharpness. 

Mirror sharpness also retrieves previous notions of curvature studied for preconditioned gradient descent \citep{CGKAM+23, CDTKL24, ICCG26}, which we show in \Cref{sec:pgd}. Furthermore, while our definition resembles the \emph{generalized sharpness} presented in \cite{ICCG26}, their work studies \emph{steepest descent} algorithms in arbitrary norm, which induce a different \emph{fixed} geometry on the space than mirror descent. In this sense, mirror sharpness can be viewed as generalized sharpness with respect to a point-wise norm. 

One benefit of mirror sharpness is that it is relatively simple to compute, as it can be expressed as a generalized eigenvalue. If we assume that $\nabla^2 \Phi(x)$ is positive-definite, then the mirror sharpness can be written as a symmetric eigenvalue problem via Rayleigh--Ritz:

\begin{lemma}
\label[lemma]{lem:rayleigh-ritz}
    Let $x \in E$ and $G = \nabla^2 \Phi (x) \succ 0$. Then, $S^\Phi(x) = \maxEV{G^{-1/2}(x)\nabla^2 L(x)G^{-1/2}(x)}.$
\end{lemma}

\subsection{Second-order pictures of mirror sharpness}

We show how, at the second order, we can retrieve familiar results in terms of the mirror sharpness.

The $\ell_2$ sharpness in ordinary gradient descent is significant because it can characterize when $L$ decreases between two iterates $x_t$ and $x_{t+1}$. In the case of mirror descent, this relationship is murkier because the geometry induced by $\Phi$ changes between $x_t$ and $x_{t+1}$. However, this asymmetry only arises at the \emph{third} order  locally:
\[D_{\Phi}(x_t, x_{t+1}) = D_\Phi(x_{t+1}, x_t) + O(\|x_{t+1} - x_t \|^3).\]
Consequently, at second order, using standard arguments, we recover a descent lemma \citep{BSTV18} with the familiar $2/\eta$ threshold.

\begin{proposition}[Second-order (approximate) descent lemma]
\label[proposition]{prop:descent}
Let $\nabla^2 \Phi (x_t) \succ 0$. Let $x_{t+1}$ be a mirror descent update with step size $\eta$. Then,
\[
    L(x_{t+1})- L(x_t)
    \leq
    \left(
        S^\Phi(x_t)-\frac{2}{\eta}
    \right)
    D_\Phi(x_{t+1},x_t)
    +
    O\left(\|x_{t+1}-x_t\|^3\right).
\]
\end{proposition}

Next, we restrict the loss to a quadratic objective, and  we show that for any mirror map $\Phi$, mirror descent can be reinterpreted as gradient descent. To our knowledge, this appears not to be a readily known result in the literature. Using this equivalence, we show that mirror sharpness naturally emerges as the sharpness of the reparameterized loss function.

\begin{theorem}[Quadratic mirror descent as gradient descent]
\label[theorem]{thm:quadratic_equivalence}
    Let $L(x) :=\frac{1}{2} \langle x - x^\star, A (x-x^\star)\rangle$ with $A \succ 0$. Let $\nabla^2 \Phi (x) \succ 0$ for all $x \in E$. Define $z^\star := \nabla \Phi(x^\star)$ and $y := A^{-1/2}(z-z^\star)$, and let
\begin{align*}
    \widehat{L}(y) := \Phi^*(z^\star+A^{1/2}y) - \Phi^*(z^\star) - \langle x^\star,A^{1/2}y\rangle
    = D_{\Phi^*}(z^\star + A^{1/2}y, z^\star ).
\end{align*}
Then, the dual mirror descent updates $z_k$ on $L$ correspond to the gradient descent updates $y_k$ on $\widehat{L}$ (i.e., $z_k = z^\star + A^{1/2}y_k$).
\end{theorem}

This equivalence shows that the quadratic picture of mirror descent in the dual is essentially the same as gradient descent. Thus, any ``new'' behavior that is different from gradient descent must arise through third-order interactions of the loss, as we later argue in \Cref{sec:sometimes_OK}. Lastly, we identify mirror sharpness with $\ell_2$ sharpness in the reparameterized coordinates:

\begin{proposition}
\label[proposition]{prop:quadratic_sharpness}
    Let $\widehat{L}$ be defined as in \Cref{thm:quadratic_equivalence} and $y \in E$. Set $x := \nabla \Phi^* (z^\star + A^{1/2} y)$ (i.e., $x$ is the corresponding coordinate of $y$) and take $\nabla^2 \Phi (x) \succ 0$. Then, $\maxEV{\nabla^2 \widehat{L}(y)} = S^\Phi(x).$
\end{proposition}

\section{Mirror sharpness stabilizes at $2/\eta$}
\label{sec:mirror_stabilizes}

In this section, we empirically demonstrate that just as sharpness hovers around $2/\eta$ in gradient descent, our notion of mirror sharpness also hovers around $2/\eta$. 

\subsection{Preconditioned gradient descent (PGD)}
\label{sec:pgd}
As a sanity check, we apply our definition of mirror sharpness to preconditioned gradient descent (PGD), which can be rewritten as mirror descent with a quadratic potential $\Phi_{\text{PGD}}(x) = \frac{1}{2} \langle x, Px\rangle$ for a matrix $P \succ 0$. In this setting, because $\nabla \Phi_{\text{PGD}}$ is linear, \Cref{eq:MD} reduces to the update $x_{t+1} = x_t - \eta P^{-1} \nabla L(x_t).$
Mirror sharpness recovers previous notions of sharpness for PGD as previously studied in \citep{CDTKL24, ICCG26}:
\[ S^{\Phi_{\text{PGD}}}(x) = \max_{\langle u, Pu \rangle = 1} \langle u, \nabla^2 L(x)u\rangle  = \max_{\|v\|_2 = 1} \langle P^{-1/2} v, \nabla^2 L(x) P^{-1/2} v \rangle,\]
where the second equality follows from a change of variables $u \mapsto P^{-1/2}v$.

\subsection{Non-Euclidean mirror potentials}
\label{sec:non-euclidean}

We demonstrate EoS behavior with various architectures trained on a subset of CIFAR-10. All experiments in this section are run with full-batch mirror descent, ReLU activations, and MSE loss. Specific architectural and implementation details can be found in \Cref{app:experimetal_details}. In general, we have found these behaviors rather consistent across several architectures, with further experiments shown in \Cref{app:more_experiments}. 

\paragraph{Hypentropy}

We consider the class of \emph{hypentropy} mirror potentials 
\[\Phi_{\text{hyp}, \gamma}(x) = \sum_{i} \left( x_i \operatorname{arcsinh}\left(\frac{x_i}{\gamma}\right) - \sqrt{x_i^2 + \gamma^2} \right),\quad \gamma > 0\] proposed by \citet{GHS19}. The mirror descent update induced by this class interpolates between a Euclidean update and the multiplicative weights update. As $\gamma$ approaches 0, hypentropy acts similarlu to entropy, and as $\gamma$ grows large, hypentropy acts as a scaled Euclidean distance.

\begin{figure}[H]
    \centering
    \includegraphics[width=1.0\linewidth]{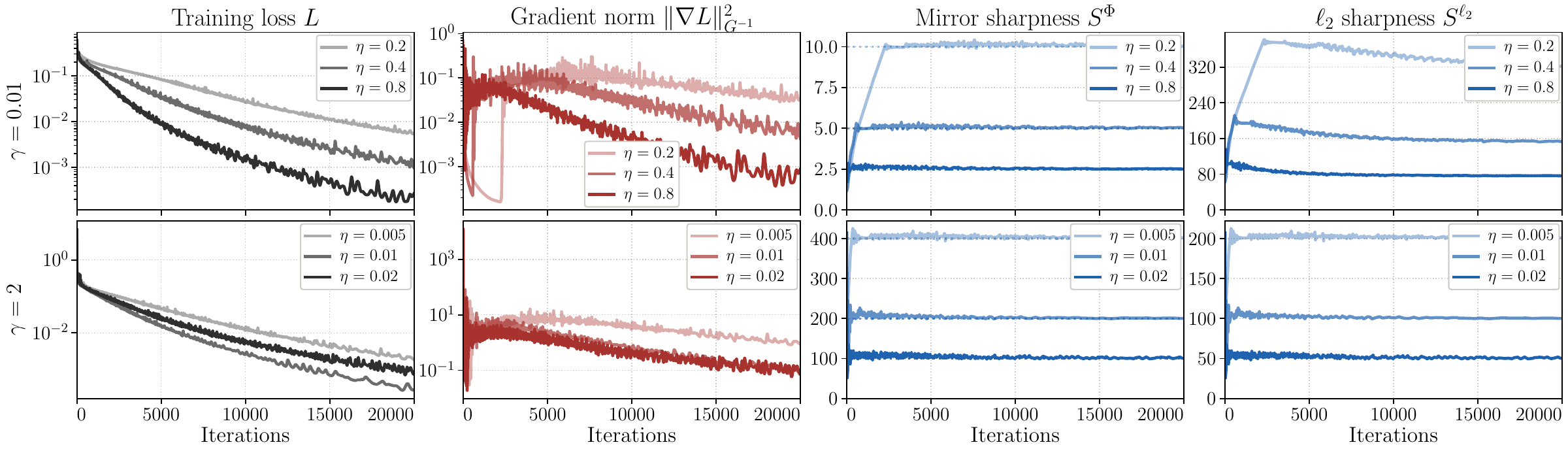}
    \caption{CNN with a hypentropy potential. Columns show training loss, norm of the loss gradient with respect to $G$, mirror sharpness, and $\ell_2$ sharpness. The $2/\eta$ thresholds are indicated by the horizontal dashed lines. Top: $\gamma = 0.01$, bottom: $\gamma = 2$.}
    \label{fig:main_text_cnn_hyp_mse}
\end{figure}

In \Cref{fig:main_text_cnn_hyp_mse}, we see that for a variety of step sizes $\eta$, the mirror sharpness $S^\Phi$ stabilizes just over $2/\eta$. This is consistent with previous findings for full-batch gradient descent in \citet{CKLKT21} and non-Euclidean steepest descent in \citet{ICCG26}. In contrast, the $\ell_2$ sharpness does not stabilize at $2/\eta$; furthermore, in the very non-Euclidean case of $\gamma=0.01$, the $\ell_2$ sharpness appears to drift substantially even after EoS.

\paragraph{$\ell_p$ norm}

We consider the class of mirror potentials given by $\Phi_{\ell_p}(x) = \frac{1}{p} \|x\|_p^p$ for $p \in (1, \infty)$, whose implicit regularization properties are studied in \citep{ALH22, SATA22, JTA26}. This class of maps has received attention because it is implementable coordinate-wise and also has been empirically observed to generalize better than ordinary gradient descent ($p = 2$) in some settings \citep{SATA22}.

\begin{figure}[H]
    \centering
    \includegraphics[width=1.0\linewidth]{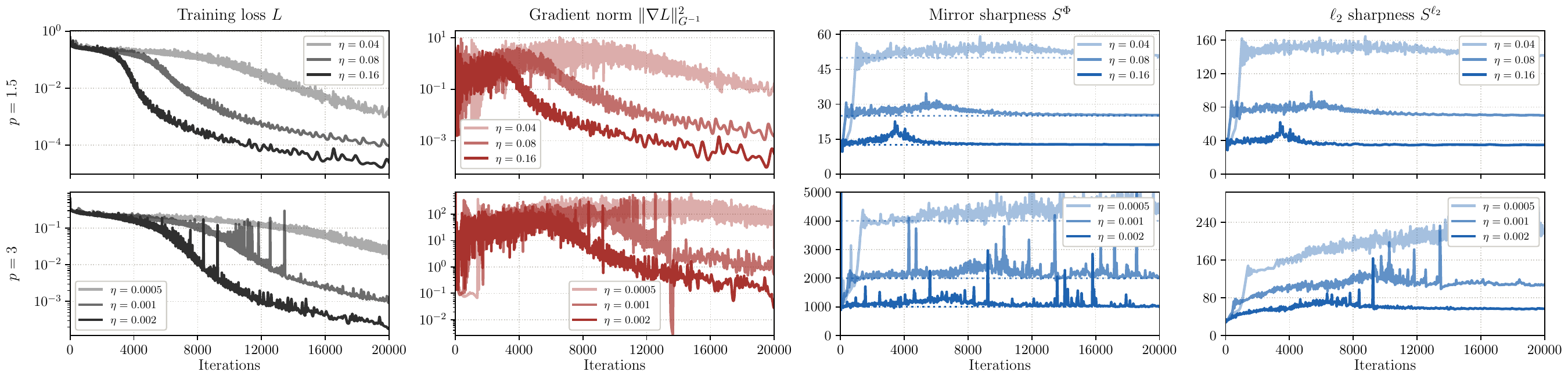}
    \caption{Vision transformer (ViT) with an $\ell_p$ potential. Columns show training loss, norm of the loss gradient with respect to $G$, mirror sharpness, and $\ell_2$ sharpness. The $2/\eta$ thresholds are indicated by the horizontal dashed lines. Top: $p = 1.5$, bottom: $p = 3$.}
    \label{fig:main_text_vit_pnorm_mse}
\end{figure}

One peculiarity of this class of maps for $p \neq 2$ is that the Hessian $\nabla^2 \Phi(x)$ becomes problematic whenever $x_i = 0$, so mirror sharpness cannot be defined on this set. In practice, this is typically not a significant issue unless parameters are zero at initialization because this region is measure zero. More specifically, for $p \in (1,2)$, $\Phi_{\ell_p}$ fails to be $C^2$ here, so $\nabla^2\Phi$ is not defined. For $p > 2$, $\nabla^2 \Phi(x)$ has a zero eigenvalue whenever some $x_i = 0$. Outside of this region, one may compute
\[[\nabla^2 \Phi_{\ell_{p}}(x)]_{ij}^{-1} = \begin{cases}
    \frac{1}{p-1}|x_i|^{2-p} & i=j \\
    0 & i \neq j.
\end{cases}\]

In general, we observe that for $p \in (1, 2)$, as $x_i \to 0$, we have $G^{-1}_{ii} \to 0$. In practice, this suppresses certain coordinates in $H$ when computing the mirror sharpness via \Cref{lem:rayleigh-ritz}. By contrast, in the $p > 2$ regime, as $x_i \to 0$, we have $G_{ii}^{-1} \to \infty$. This \emph{amplifies} certain coordinates when computing mirror sharpness, and thus  the mirror sharpness sharply spikes when some $x_i$ is close to zero. In our experiments, these effects are most significant near initialization. 

The ill-posed nature of $\nabla^2 \Phi$ does not appear to significantly empirically alter EoS dynamics, and our theoretical results remain valid away from the problematic set. We find these empirical results very interesting, and we believe that they may hint that some more ``universal'' mechanism is at play that manifests in the EoS phenomenon.

\subsection{Mirror sharpness and ordinary sharpness}

We discuss the behavior of the well-studied $\ell_2$ sharpness, whose connection to generalization has been extensively studied, such as in \citep{KCL23, ACMFH23}. Furthermore, \citet{CGKAM+23} empirically showed that while Adam maintains a \emph{preconditioned} sharpness\footnote{This notion of sharpness is the same as discussed in \Cref{sec:pgd}, only that $P$ may vary with time.}, Adam has the capability to arbitrarily increase the sharpness. This fact appears connected to how the Frobenius norm of the preconditioner matrix $P_t$ is observed to \emph{increase} throughout training time\footnote{This is, however, not universally true for any arbitrary loss, as there exist convex optimization problems such that this preconditioner norm does not monotonically increase \citep{RKK18}.}, allowing $\maxEV{H}$ to increase arbitrarily while fixing $\maxEV{P_t^{-1} H}$ around Adam's corresponding stability threshold.

In the case of mirror descent, can we predict $S^{\ell_2}$ at EoS? In the most general sense where we can take any $\Phi$, the answer is no: taking $\Phi = \frac{1}{2c} \|x\|_2^2$ yields $S^{\Phi} = cS^{\ell_2}$. Fixing a mirror map $\Phi$, however, we can make the following naive estimate of the ratio $S^{\ell_2}/S^{\Phi}$:
\begin{proposition}[Sharpness ratio]
    \label[proposition]{prop:ord_sharpness_bounds}
    Assume $S^\Phi(x) > 0$. Then, \[\minEV{\nabla^2\Phi(x)}\leq {S^{\ell_2}(x)}/{S^\Phi(x)} \leq \maxEV{\nabla^2\Phi(x)}.\]
\end{proposition}
For some mirror maps, we can bound $\maxEV{\nabla^2\Phi}$. For instance, in the case of the hypentropy potential $\Phi_{\text{hyp}, \gamma}$, we can compute $\maxEV{\nabla^2\Phi(x)} \leq \frac{1}\gamma$; we empirically observe this in all of our experiments. In these cases, if we assume that $S^\Phi$ is approximately $2/\eta$, then this implies that $S^{\ell_2}$ \emph{cannot} arbitrarily increase-- in contrast to Adam. In fact, as mentioned previously, in certain runs (e.g, \Cref{fig:main_text_cnn_hyp_mse}), we found that $S^{\ell_2}$ may \emph{decrease} after the onset of EoS. In light of \Cref{prop:ord_sharpness_bounds}, we find it surprising that, in our experiments, the ratio $S^{\ell_2}/S^\Phi$ remains relatively stable and occupies a very narrow range within what is predicted by \Cref{prop:ord_sharpness_bounds}.

\section{A dual central flow for mirror descent}
\label{app:mirror_flow}

In the remainder of this paper, we attempt to study mirror descent through the \emph{central flow} framework \citep{CDTKL24}, which is an ODE that describes a \emph{time-averaged} (i.e., locally smooth) trajectory $\bar{x}(t)$ of iterates at EoS. The Riemannian gradient flow fails to be a good approximation at EoS because it does not fix the sharpness at around $2/\eta$; thus, the Riemannian gradient flow significantly diverges from the mirror descent iterates after the onset of EoS. We show this empirically in \Cref{sec:empirical_flow}. 

While this ODE may be of independent interest as a more tractable object to study than mirror descent, our main goal is to understand the extent to which the self-stabilization theory holds in the non-Euclidean setting of mirror descent, so we use this picture as an ansatz for our central flow. We find that such a framework is not necessarily well posed without adding extra constraints on the \emph{third-order} interactions between the loss and $\Phi$; such a result is interesting because it raises a challenge in extending the self-stabilization argument to mirror descent \citep{ALP22, DNL23, CDTKL24, LKK26}. 

\subsection{Deriving the central flow}

\textbf{We will assume that $L, \Phi$ are $C^4$.} We will restrict our central flow to the case when only one eigenvalue of $H$ is at (approximately) $2/\eta$. Let $u=u(\bar{x})$ be the top eigenvector of $(H,G)$ that solves the maximization problem defined in \Cref{def:sharpness}. Even in this case, there are complexities caused by the non-Euclidean nature of mirror descent.

The main idea behind central flows is modeling the iterates $x_t$ as an oscillation about a \emph{time-averaged}\footnote{This expectation represents a smoothening/averaging of $\xi(t)r(t)$, though this is left deliberately vague because it is not easy to tell what time scale to average the terms with. We lift the same assumptions about the expectation operator $\mathbb{E}[\cdot]$ from \citet[Appendix A.1.5]{CDTKL24}, where one may find a more detailed discussion.} $\bar{x}(t)$ along the unstable eigenvector $u(t)$. More concretely, \citet{CDTKL24} models $x_t = \bar{x}(t) + \xi(t) u(t)$.\footnote{Even in the case of gradient descent, such a model is a simplification, as $x_t - \mathbb{E}[x_t]$ does not necessarily have to lie in the span of $u(t)$.} It is unclear to what extent such a phenomenon holds for the case of mirror descent \emph{especially} because neither the primal nor the dual space is implicitly endowed with an $\ell_2$ geometry in general. To make progress towards answering this question, let us naively assume that such a model is valid.

Specifically, we model oscillations in the \emph{dual} space via $z_t = \bar{z}(t) + \xi(t) r(t)$, where $r = Gu$ is the pushforward of $u$ under the mirror map $\nabla \Phi$ (i.e., the natural way tangent vectors in the primal space are mapped to tangent vectors in the dual space).\footnote{More formally, we view $u$ as an element of the tangent space $T_xE$. We can rewrite the differential $D(\nabla\Phi)_x = \nabla^2\Phi(x)=G$, so $r = Gu$ is the push-forward of $u$ and lives in $T_{\nabla \Phi(x)}E^*$.} Furthermore, we impose $\mathbb{E}[\xi(t)] = 0$ and $\mathbb{E}[\xi(t)r(t)] \approx \mathbb{E}[\xi(t)] r(t) = 0$. We believe that the dual is a more natural choice to model than the primal because the primal update introduces additional non-linearities via the mirror map. Recall that the dual update is given by:
\[z_{t+1} - z_{t} =  - \eta (\nabla L \circ \nabla \Phi^*)(z_t).\]
We abbreviate $F:= \nabla L \circ \nabla \Phi^*$. With these assumptions and regularity where appropriate, approximating $d\bar{z}/dt \approx \bar{z}_{t+1} - \bar{z}_{t}$ and expanding all relevant terms to $O(\xi^3)$ yields the \emph{dual central flow} ODE:
\begin{equation}
 \label{eq:DCF}
    \frac{d\bar{z}}{dt} = -\eta \left( F(\bar{z}) + \frac{\sigma^2}{2} D^2 F(\bar{z})[r,r]\right),\tag{DCF}   
\end{equation}
where $\sigma^2 := \mathbb{E}[\xi(t)^2]$. By further constraining $\widetilde{S}^\Phi(\bar{z}) := ( S^\Phi \circ \nabla \Phi^*)(\bar{z}) = 2/\eta$ to model iterates at EoS, we can solve for a unique value of $\sigma^2$ to pin down the trajectory: 
\[\sigma^2 = \frac{-2\langle \nabla \widetilde{S}^\Phi(\bar{z}), F(\bar{z})\rangle}{\langle \nabla \widetilde{S}^\Phi(\bar{z}), D^2F(\bar{z})[r,r]] \rangle}.\]
As much of the details of this derivation are similar to those in \cite{CDTKL24}, we defer a complete derivation to \Cref{app:central_flow_derivation}. We observe that in the case where $\Phi = \frac{1}{2} \|x\|^2_2$, we retrieve the central flow for ordinary gradient descent in \cite{CDTKL24}.

Finally, we retrieve a flow for the primal quantity $\bar{x}(t) := \nabla \Phi^*(\bar{z}(t))$. Again, as $d\bar{x}/dt$ is a tangent vector on the primal space, we use the pullback $G^{-1}$ to relate it with $d\bar{z}/dt$:
\begin{align*}
 \label{eq:PCF}
    \frac{d\bar{x}}{dt} &=  G^{-1} \frac{d \bar{z}}{dt} = -\eta G^{-1}\left( \nabla L(\bar{x}) + \frac{\sigma^2}{2}\left(\nabla^3L(\bar x)[u,u] - HG^{-1}\nabla^3 \Phi(\bar x) [u,u] \right)\right)
    \tag{PCF}   
\end{align*}
We call this the \emph{primal central flow}.\footnote{Note however, that $\bar{x} \neq \mathbb{E}[x_t]$ in general due to the non-linearities of $\nabla \Phi^*$.} In the second equality, we re-express the dual quantities in \Cref{eq:DCF} in primal quantities.

\subsection{Empirical verification}
\label{sec:empirical_flow}

We verify empirically that \Cref{eq:DCF} describes iterates at the edge of stability more accurately than the Riemannian gradient flow in the dual space, which is given by
\begin{equation}
    \label{eq:RGF}
    \frac{d\bar{z}}{dt}  = -\eta\nabla L(\bar{x}).
    \tag{RGF}
\end{equation}
For comparison, we also plot the trajectory of gradient flow (i.e., where we let $G$ be the identity) to show that it diverges sharply from the mirror descent trajectory. 
\begin{figure}[H]
    \centering
    \includegraphics[width=0.8\linewidth]{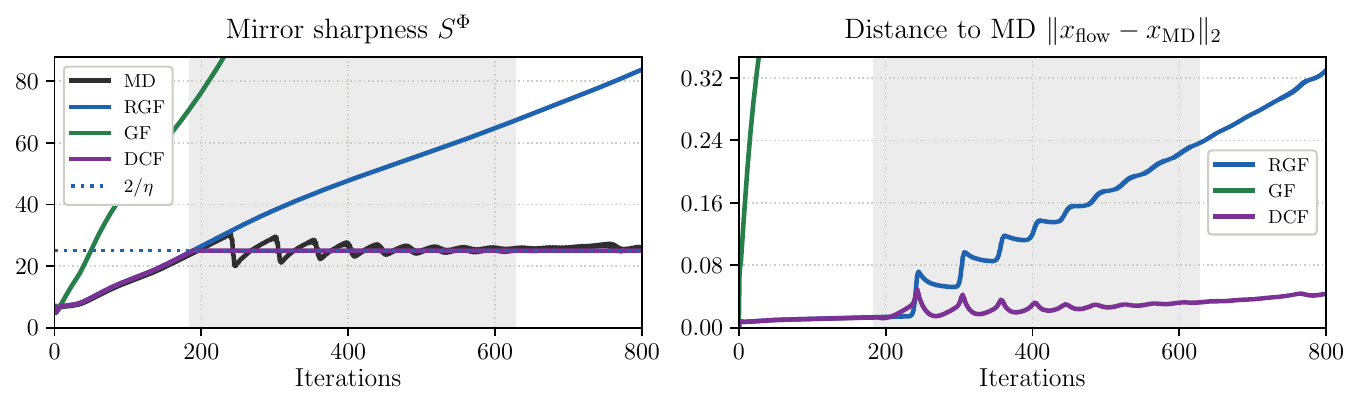}
    \caption{CNN with an $\ell_{1.5}$ potential, GeLU activations, MSE loss, and $\eta = 0.08$ on a subset of CIFAR-10. Along with the mirror descent iterates (MD), we track the Riemannian gradient flow (RGF), gradient flow (GF), and our dual central flow (DCF). We indicate, in gray, the timesteps where the DCF iterates only have one unstable eigenvector with eigenvalue greater than $1.95/\eta$. Left: Mirror sharpness. Right: Distance to mirror descent iterates.}
    \label{fig:main_text_central_flows}
\end{figure}
We find that the dual central flow tracks the mirror descent iterates fairly well at EoS, improving over the Riemannian gradient flow, even when there is more than one unstable eigenvalue. Furthermore, our dual central flow straddles the $2/\eta$ threshold for mirror sharpness very closely. We also remark that discretizing \Cref{eq:DCF} is non-trivial since we require that the mirror sharpness is constrained \emph{exactly} at $2/\eta$. We comment more on this in \Cref{app:discrete}. Further experiments can be found in \Cref{app:ce_experiments}.

\subsection{The central flow (may) encode self-stabilization}
\label{sec:sometimes_OK}

Let us make more explicit how the central flow \Cref{eq:DCF} ``encodes'' the past result of \citet{DNL23} explaining why $\ell_2$ sharpness stabilizes at $2/\eta$ in the case of gradient descent. Assuming a well-separated gap between the first and second eigenvalues, one can solve explicitly for $\frac{d}{dt} S^\Phi$:

\begin{proposition}
    \label[proposition]{prop:sharpness_time_derivative}
    Let $\bar{z}$ evolve according to \Cref{eq:DCF}. Let $L, \Phi$ be $C^4$ and $\nabla^2 \Phi \succ 0$ for all $x$ along $\bar{x}$. Then,
    \[\frac{d}{dt} \widetilde{S}^\Phi(\bar{z}) = \eta \langle \nabla \widetilde{S}(\bar{z}), - F(\bar{z}) \rangle - \frac{\eta \sigma^2}{2} \langle \nabla \widetilde{S}(\bar{z}), D^2F(\bar{z})[r,r] \rangle,\]
    where $r = Gu$ and $u$ is the top eigenvector of $(H,G)$.
\end{proposition}

The first term admits a very simple interpretation as the \emph{progressive sharpening coefficient}. In dual space, the gradient update is given by $-F(z)$, so $\widetilde{S}$ increases if and only if $\langle \nabla \widetilde{S}(z), -F(z) \rangle > 0$. 

\begin{definition}
    \label[definition]{def:PSCoeff}
    The \emph{progressive sharpening coefficient} is given by
    \[\alpha^{\Phi}(z) := -\langle \nabla \widetilde{S}(z), F(z) \rangle.\]
\end{definition}

In the case of gradient descent, $\alpha^\Phi$ is assumed to be positive by \citet{DNL23} and verified empirically. 

To counteract the positive contribution to $S^\Phi$, \citet{DNL23} explains that the oscillations along the top eigenvector ``self-stabilize'' the sharpness. In the context of gradient descent, the strength of this restoring force is quantified by $\beta^{\ell_2}(x) = \|\nabla S^{\ell_2}(x)\|^2_2 \geq 0$. As we may expect, the second term in \Cref{prop:sharpness_time_derivative} reveals the mirror descent analogue of $\beta^{\ell_2}$, as $D^2 F(\bar{z})[r,r]$ reduces to $\nabla S(x)$ in the case of gradient descent.

\begin{definition}
    \label[definition]{def:stabCoeff}
    The \emph{mirror stabilization coefficient} $\beta^{\Phi}(z)$ is given by:
    \[\beta^\Phi(\bar z) := \langle \nabla \widetilde{S}(\bar{z}), D^2F(\bar{z})[r,r] \rangle,\]
    where $r = Gu$ and $u$ is the top eigenvector of $(H,G)$. Equivalently, by \Cref{prop:primal_beta},
    \[\stabCoeff^\Phi(\nabla \Phi(x)) = \|\nabla S\|^2_{G^{-1}} + \langle G^{-1} \nabla S, (SI - HG^{-1}) \nabla^3 \Phi[u,u] \rangle.\]
\end{definition}
Using the primal expansion of $\beta^\Phi$, we observe that for quadratic potentials, $\beta^\Phi \geq 0$. However, this need not be true in general; we show the ``geometry correction'' term can be made large, pushing $\beta^\Phi$ to be negative:

\begin{lemma}[$\stabCoeff^\Phi$ can be negative]
    \label[lemma]{lem:negative_beta}
    Fix $x^\star \in E$ where the ambient dimension $d > 1$, and some mirror map $\Phi$ with $\nabla^3 \Phi(x^\star) \neq 0$. There exists some $L$ with bounded $\|\nabla^3L[\cdot, \cdot, \cdot]\|$ such that $\stabCoeff^\Phi(x^\star) < 0$ and $S^\Phi(x^\star)$ can be made any positive number (e.g., $2/\eta$). 
\end{lemma}

Why is this problematic? If we assume that progressive sharpening happens with $\alpha^\Phi > 0$, having $\beta^\Phi < 0$ \emph{forces} $\widetilde{S}^\Phi (\bar{z})$ to increase via \Cref{prop:sharpness_time_derivative}. Thus, in this case, when we solve for the value of $\sigma^2 = 2\alpha^\Phi/\beta^\Phi$, we would find that $\sigma^2$, which represents the variance of the oscillations, \emph{is negative} and thus the central flow fails to be physically relevant. We conclude that the universal self-stabilization mechanism via oscillations is not currently sufficient to explain EoS in mirror descent. 

While our result, as stated, concerns only a single point $x^\star$, we believe that it is generally possible to use this construction to generate explicit examples of optimization around $x^\star$ such that $\beta^\Phi < 0$ along a time-averaged trajectory for \emph{any} (locally nonquadratic) mirror map. We have provided such an example in $\mathbb{R}^2$ via a cubic loss in \Cref{app:numerical_example} for the hypentropy map, where $\alpha^\Phi > 0$ but $\beta^\Phi <  0$ along an optimization trajectory slightly above the edge of stability.

In light of this negative result, we believe it is quite difficult, given the current theory, to argue for a self-stabilization result without enforcing substantial conditions that are not easy to interpret. While it may be possible to sidestep the obstacle mentioned in \Cref{lem:negative_beta} by arguing that the central flow is repelled from regions where $\beta^\Phi < 0$, we find this difficult because even arguing why $\alpha^\Phi > 0$ holds is a major open question for ordinary gradient descent. In our neural-network experiments in \Cref{app:ce_experiments}, we plot values of $\alpha^\Phi, \beta^\Phi$ along the dual central flow trajectory, and we find that $\beta^\Phi > 0$ always.

\section{Conclusion and future work}

In this work, we studied edge of stability behavior in the non-Euclidean setting of mirror descent, by introducing a mirror sharpness that recovers the familiar $2/\eta$ threshold for the edge of stability. Empirically, we show that the EoS phase manifests through various step sizes, architectures, and mirror maps. However, through the framework of central flows, we identify that a ``self-stabilization'' picture in the dual space does not sufficiently explain why mirror descent should stabilize due to third-order interactions between the mirror potential and the loss.

It remains unclear if mirror descent ``universally'' self-stabilizes, even if we assume progressive sharpening, in realistic losses. Towards a positive answer, one could argue that the situations in \Cref{lem:negative_beta} where $\beta^\Phi < 0$ are \emph{naturally avoided} by mirror descent, or, alternatively that points where $\beta^\Phi < 0$ do not naturally show up in neural network losses. Towards a negative answer, one may try to exhibit a realistic example of successful optimization where $\beta^\Phi(z) < 0$ and the mirror sharpness climbs arbitrarily high beyond the edge of stability.

\subsection*{AI use statement}

Almost all of our code was generated by generative AI tools, although the first author has carefully checked the output and takes full responsibility. We plan to release code at a slightly later date. We have also used generative AI tools to do some technical computations. The authors take full responsibility for the correctness of this work.

\subsection*{Acknowledgments}

We acknowledge the computational resources provided by the Aalto Science-IT project. We would also like to thank several anonymous reviewers whose comments have improved the paper.

\bibliography{references}
\bibliographystyle{alpha}

\appendix

\section{More on the central flow}

\subsection{Central flow derivation}
\label{app:central_flow_derivation}

We will consider the appropriate regularity $L, \Phi \in C^4$. We will work in the dual coordinates $z = \nabla \Phi (\bar{x})$. Recalling $F = \nabla L\circ \nabla \Phi^*$, we may write the dual mirror descent update:
\[z_{k+1} = z_k - \eta F(z_k),\]

We model iterates $z_k$ as oscillating along the direction $r$, which is the pushforward of $u$ through $\nabla \Phi$, given by $r = Gu$. Then, $r \in E^*$ satisfies $\langle r, \nabla^2 \Phi^*(z)r\rangle = 1$, and we model $z(t) := \bar{z}(t) + \xi(t) r(t)$ with $\mathbb{E}[\xi(t)] = 0$.  Let $\sigma^2(t) := \mathbb{E}[\xi(t)^2]$.
One may Taylor expand $F$ at $\bar{z}$ along the $r$ direction:
\[F(\bar{z} + \xi r) = F(\bar{z}) + \xi DF(\bar{z})[r]\ + \frac{\xi^2}{2}D^2F(\bar{z})[r,r] + O(\xi^3).\]
After averaging, we have:
\begin{align*}
   \avg{F(\bar{z} + \xi r)} &= \avg{F(\bar{z}) + \xi DF(\bar{z})[r]\ + \frac{\xi^2}{2}D^2F(\bar{z})[r,r]} + O(\avg{|\xi|^3}) \\
   &\approx F(\bar{z}) + \frac{\sigma^2}{2}D^2F(\bar{z})[r,r] + O(\avg{|\xi|^3}).
\end{align*}
Here, we take some care in handling the second term 
\[\mathbb{E}[\xi DF(\bar{z})[r]] = DF(\bar{z}) [\mathbb{E}[\xi r]] \approx DF(\bar{z}) [\mathbb{E}[\xi]r] = 0,\]
as $DF(\bar{z})$ acts linearly on $r$.
At the time $t$ that corresponds to an iterate $z_k$, we may approximate
\[\frac{d\bar{z}}{dt} \approx \bar{z}_{k+1} - \bar{z}_{k} = -\eta \mathbb{E} [F(\bar{z}_k + \xi r)].\]
Combining the two equations yields the dual central flow ODE:
\[\frac{d\bar{z}}{dt} = -\eta \left( F(\bar{z}) + \frac{\sigma^2}{2} D^2 F(\bar{z})[r,r]\right).\]
Define $\widetilde{S} = S \circ \nabla\Phi^*$. 

In the continuous-time limit, to solve for $\sigma^2(t)$, we may simply assume that the iterates $\bar{z}$ are already at the edge of stability:
\[\widetilde{S}(\bar{z}(t)) = \frac{2}{\eta},\]
and therefore to ensure that $\bar{z}$ stays here, we enforce
\[\frac{d}{dt}(S \circ \nabla \Phi^*(\bar{z})) = 0\]
Then, by the chain rule, the LHS can be re-expressed as $\langle \nabla \widetilde{S}, \frac{d\bar{z}}{dt} \rangle$, which implies that
\[ \left\langle \nabla \widetilde{S}(\bar{z}), -\eta \left( F(\bar{z}) + \frac{\sigma^2}{2} D^2 F(\bar{z})[r,r]\right) \right\rangle = 0.\]
Solving for $\sigma^2$ yields:
\[\sigma^2 = \frac{-2\langle \nabla \widetilde{S}(\bar{z}), F(\bar{z})\rangle}{\langle \nabla \widetilde{S}(\bar{z}), D^2F(\bar{z})[r,r]] \rangle}.\]
Observe that the numerator is exactly $2\PSCoeff^{\Phi}$ as $\nabla \widetilde{S}(\bar{z}) = G^{-1} \nabla S(\bar{x})$ by~\Cref{lem:dual_gradient} and the denominator is equal to $\stabCoeff^\Phi$ by~\Cref{prop:primal_beta}. To retrieve \Cref{eq:PCF}, we directly apply \Cref{prop:curv2curv}.

\subsection{Discretizing the central flow}
\label{app:discrete}

In practice, however, we cannot assume that $\widetilde{S}(\bar{z}(t))$ will rise continuously until it hits $2/\eta$ due to discretization effects. Let us suppose that the top eigenvalue of $(H,G)$ is close to, but not over, $2/\eta$.

Define the sharpness residue $A(z) := 2/\eta - \widetilde{S}(z)$. Let us write down the following \emph{differential complementarity problem} (DCP) in $\zeta$:
\[\frac{d\bar{z}}{dt} = -\eta \left( F(\bar{z}) + \frac{\zeta}{2} D^2 F(\bar{z})[r,r]\right), \quad\quad \text{where} \quad 0 \preceq \zeta \perp A(\bar{z}) \succeq 0.\]
We use $\zeta \perp A(\bar{z})$ (where $\perp$ denotes complementarity) to enforce the condition that $\zeta A(\bar{z}) = 0$.

Discretizing this DCP is discussed in good detail in \citet[Appendix A.6.1]{CDTKL24}, and we adapt their method. While this is not practical in general in the case of multiple unstable eigenvectors as \citet{CDTKL24} describes, for our purposes of modeling one unstable eigenvector, it is satisfactory. Define the linearization of $A$:
\begin{align*}
   A^{\text{lin}}(z) &:= A(\bar{z}) + \langle \nabla A( \bar{z}), z - \bar{z} \rangle \\
   &= 2/\eta - \widetilde{S}(\bar{z}) - \langle \nabla \widetilde{S}( \bar{z}), z - \bar{z} \rangle 
\end{align*}
Let $t'$ denote the current time step, and $t' + \epsilon$ denote the next timestep. We would like to enforce the DCP at $\bar{z}^{(t' + \epsilon)} = \bar{z}^{(t')} + \epsilon \frac{d\bar{z}^{(t')}}{dt}$. Thus,
\begin{align*}
    A^{\text{lin}}(\bar{z}^{(t' + \epsilon)}) &= 2/\eta - \widetilde{S}(\bar{z}^{(t')}) - \left\langle \nabla \widetilde{S}( \bar{z}^{(t')}), \epsilon \frac{d\bar{z}^{(t')}}{dt} \right\rangle\\
    &=2/\eta - \widetilde{S}(\bar{z}^{(t')}) -\eta \epsilon \alpha^\Phi(\bar{z}^{(t')}) + \frac{\eta\epsilon\zeta}{2} \beta^\Phi(\bar{z}^{(t')}),
\end{align*}
where the last equality follows from \Cref{prop:sharpness_time_derivative}. Thus, enforcing the complementarity condition at $\bar{z}^{(t' + \epsilon)}$ yields
\[\zeta = 0 \text{ or } \zeta = \frac{2}{\beta^\Phi(z^{(t')})} \left( \alpha^\Phi(z^{(t')}) - \frac{ 2/\eta-\widetilde{S}(z^{(t')})}{\eta \epsilon}\right).\]
Enforcing the complementarity and $\zeta \geq 0$ and $A^{\text{lin}}(\bar{z}^{(t' + \epsilon)}) \geq 0$ yields (and assuming $\beta^\Phi(z^{(t')}) > 0$):
\[\zeta = \max\left(0, \frac{2}{\beta^\Phi(z^{(t')})} \left( \alpha^\Phi(z^{(t')}) - \frac{ 2/\eta - \widetilde{S}(z^{(t')})}{\eta \epsilon}\right) \right).\]
This becomes our choice of $\zeta$, and we implement the Euler timestep
\[\bar{z}^{(t' + \epsilon)} \gets \bar{z}^{(t')} - \eta \epsilon \left( F(\bar{z}^{(t')}) + \frac{\zeta}{2} D^2 F(\bar{z}^{(t')})[r^{(t')},r^{(t')}]\right).\]

\subsection{Comparison against \citet{CDTKL24}}

We compare our model against \citet{CDTKL24}'s central flow model for gradient descent. Surprisingly, even though we only account for the top eigenvector, our model provides very similar results. We hypothesize that this might be because we run $4$ timesteps per mirror descent iteration, so our solver can ``take turns'' and push down each eigenvector that has eigenvalue greater than $2/\eta$.

\begin{figure}[H]
    \centering
    \includegraphics[width=1.0\linewidth]{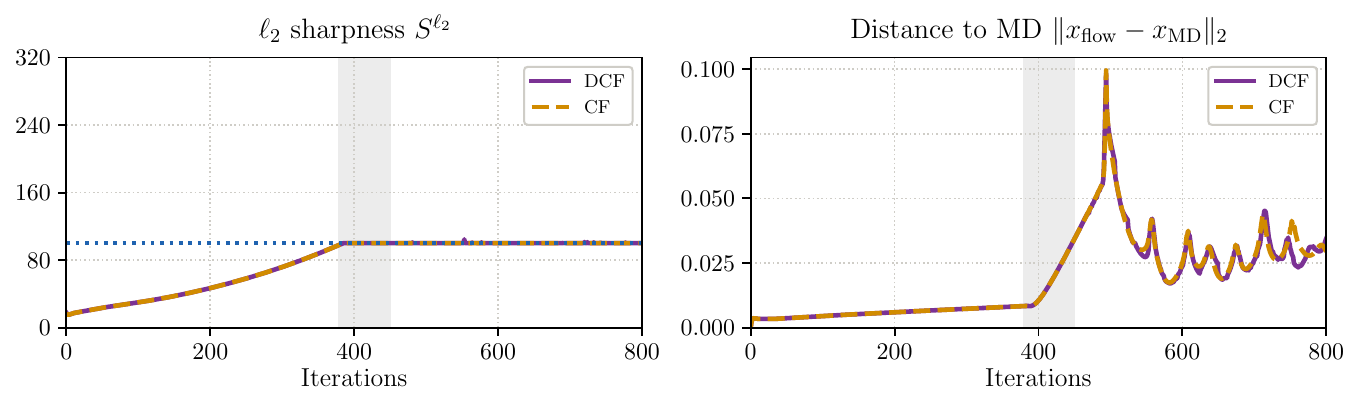}
    \caption{MLP with an MSE loss on a subset of CIFAR-10; gradient descent; $\eta = 0.02$. We compare our dual central flow (DCF) against the central flow (CF) of \citet{CDTKL24}. We indicate, in gray, the timesteps where the DCF iterates only have one unstable eigenvector with eigenvalue greater than $1.95/\eta$.}
    \label{fig:CNN_l1.5_MSE_CF}
\end{figure}

\section{A spectral interpretation of the stabilization coefficient}
\label[appendix]{sec:spectral_beta}

\begin{proposition}[Expansion in the generalized eigenbasis]
\label[proposition]{prop:spectral}
Let $u_i$ be an orthonormal basis of generalized eigenvectors satisfying
\[
H u_i = \lambda_i G u_i,
\qquad
\langle u_i, G u_j\rangle = \delta_{ij},
\]
with eigenvalues ordered so that $\lambda_1 = S$. Then,
\[
\stabCoeff^\Phi(x) = \|\nabla S^\Phi(x)\|_{G^{-1}}^2 + \sum_i (S^\Phi-\lambda_i)
\langle \nabla S^\Phi, u_i\rangle
\langle \nabla^3\Phi[u,u],u_i\rangle.
\]
\end{proposition}

While we can guarantee $\|\nabla S^\Phi(x)\|_{G^{-1}}^2$ is never negative, it appears difficult to control the geometry correction caused by the alignment of $\nabla S$ and $\nabla^3 \Phi[u,u]$. Interestingly, the eigenvalues at the edge of stability do not contribute to the geometry correction, and \emph{smaller} eigenvalues are weighted heavier in the geometry correction. This suggests that understanding EoS dynamics in mirror descent may require tracking the evolution of the entire spectrum during training, rather than only the largest eigenvalue.

\section{Deferred proofs and some identities}
\label{app:proofs-mirror-sharpness}

\subsection{Proof of \Cref{lem:rayleigh-ritz}}

    Via a homogeneity argument,
    \begin{align*}
        S(x) &= \max_{\langle u, Gu \rangle = 1} \langle u, Hu\rangle  \\
        &= \max_{u \neq 0} \frac{ \langle u, H u \rangle }{\langle u, G u\rangle}.
    \end{align*}
    Since $G\succ 0$, let $v := G^{1/2}u$. Then, one can rewrite $\langle u,Gu\rangle = \langle v,v\rangle$ and $\langle u,Hu\rangle = \left\langle v, G^{-1/2}HG^{-1/2}v \right\rangle.$
    Therefore,
    \[
        S(x) = \max_{v\neq 0}
        \frac{ \left\langle v,G^{-1/2}HG^{-1/2}v\right\rangle}{
            \langle v,v\rangle
        }.
    \]
    Since $G^{-1/2}HG^{-1/2}$ is symmetric, Rayleigh--Ritz
    gives
    \[
        S^\Phi(x) = \maxEV{G^{-1/2}HG^{-1/2}}.
    \]
    This shows the first claim. Observe that the top eigenvector $v_1$ solves
    \[S^\Phi v_1 = G^{-1/2} H G^{-1/2}v_1.\]
    Let $u_1 = G^{-1/2}v_1$, and rewrite
    \[S^\Phi G^{1/2}u_1 =G^{-1/2}Hu_1.\]
    As $G \succ 0$, we find that $S Gu_1 = Hu_1$, showing that mirror sharpness is the top generalized eigenvalue of $(H,G)$.

\subsection{Proof of \Cref{prop:descent}}

Let $\Delta_t:=x_{t+1}-x_t.$ Taylor expanding $L$ gives
\[
    L(x_{t+1})- L(x_t)
    =
    \langle \nabla L(x_t),\Delta_t\rangle + \frac{1}{2} \langle \Delta_t,H(x_t)\Delta_t\rangle +  O(\|\Delta_t\|^3),
\]
and we will bound the terms on the RHS. Writing the \Cref{eq:MD} update gives:
\[
    \nabla \Phi(x_{t+1})-\nabla \Phi(x_t)
    = -\eta \nabla L(x_t).
\]
Taking the inner product with $\Delta_t$, we obtain
\[
    \langle \nabla  L(x_t),\Delta_t\rangle
    = -\frac{1}{\eta}
    \left(
        D_\Phi(x_{t+1},x_t) + D_\Phi(x_t,x_{t+1})
    \right).
\]
Note that $D_\Phi(x_{t+1}, x_t)$ and $D_\Phi(x_{t}, x_{t+1})$ have the same Taylor expansion up to order two:
\begin{align*}
    D_\Phi(x_{t+1},x_t) &= \frac{1}{2} \langle \Delta_t,G(x_t)\Delta_t\rangle + O(\|\Delta_t\|^3), \\
    D_\Phi(x_t,x_{t+1}) &= \frac12 \langle \Delta_t,G(x_t)\Delta_t\rangle + O(\|\Delta_t\|^3).
\end{align*}
Thus, one can rewrite:
\[
    \langle \nabla  L(x_t),\Delta_t\rangle
    = -\frac{2}{\eta} D_\Phi(x_{t+1},x_t) + O(\|\Delta_t\|^3).
\]
By the definition of mirror sharpness,
\[
    \langle \Delta_t,H(x_t)\Delta_t\rangle
    \leq S(x_t) \langle \Delta_t,G(x_t)\Delta_t\rangle.
\]
To compute $\langle \Delta_t, G(x_t) \Delta_t \rangle$, we can Taylor expand $\Phi(x_{t+1})$ about $x_t$:
\[\Phi(x_{t+1}) = \Phi(x_t) + \langle \nabla \Phi(x_t), \Delta_t \rangle + \frac{1}{2} \langle \Delta_t, G(x_t) \Delta_t \rangle + O(\|\Delta_t \|^3).\]
Rearranging and substituting the Bregman divergence $D_\Phi(x_{t+1}, x_t)$ yields:
\[ \frac{1}{2} \langle \Delta_t, G(x_t)\Delta_t \rangle  = D_\Phi(x_{t+1}, x_t) + O(\|\Delta_t \|^3).\]
Combining these estimates yields the result.

\subsection{Proof of \Cref{thm:quadratic_equivalence}}

Write $z_t:=\nabla\Phi(x_t)$. Since $\nabla L(x)=A(x-x^\star),$
the mirror descent update in dual coordinates is
\[ z_{t+1} = z_t-\eta A(x_t-x^\star) = z_t-\eta A\bigl(\nabla\Phi^*(z_t)-\nabla\Phi^*(z^\star)\bigr). \]
The key observation is that, because the gradient of the quadratic objective is linear, this vector field becomes a Euclidean gradient after a linear change of coordinates. Define
\[y_t:=A^{-1/2}(z_t-z^\star), \qquad\text{such that}\qquad z_t=z^\star+A^{1/2}y_t.\]
Multiplying the dual update by $A^{-1/2}$ gives
\[
\begin{aligned}
y_{t+1} &= y_t - \eta A^{1/2} \bigl( \nabla\Phi^*(z^\star+A^{1/2}y_t)-x^\star \bigr).
\end{aligned}
\]
On the other hand, differentiating $\widehat{ L}$ yields
\[\nabla\widehat{ L}(y) = A^{1/2} \bigl( \nabla\Phi^*(z^\star+A^{1/2}y)-x^\star \bigr).\]
Thus,
\[ y_{t+1} = y_t-\eta\nabla\widehat{ L}(y_t). \]

\subsection{Proof of \Cref{prop:quadratic_sharpness}}

Differentiating $\widehat{L}$ twice yields
\[
\nabla^2 \widehat{L}(y)
= A^{1/2} \nabla^2 \Phi^*(z^\star + A^{1/2}y) A^{1/2}.\]
Note that $\nabla^2 \Phi^*(z) = (\nabla^2 \Phi(x))^{-1} = G^{-1}$, which follows from differentiating both sides of $\nabla \Phi^*(\nabla \Phi(x)) = x$. Thus, we may rewrite this as
\[
\nabla^2 \widehat{L}(y) = A^{1/2} G^{-1} A^{1/2} 
\]
We claim that the eigenvectors $\tilde{v}_i$ of $\nabla^2 \widehat{L}(y)$ are in bijection with the generalized eigenvectors $v_i$ of $(H,G)$ via $\tilde{v}_i \mapsto A^{1/2}v_i$; furthermore, their eigenvalues are preserved. Indeed,
\[
A^{1/2}G^{-1}A^{1/2}\tilde v_i=\lambda_i\tilde v_i
\quad\Longleftrightarrow\quad
G^{-1}Av_i=\lambda_i v_i
\quad\Longleftrightarrow\quad
Av_i=\lambda_iGv_i.
\]
This proves the claim by taking the top eigenvalue.

\subsection{Proof of \Cref{prop:ord_sharpness_bounds}}

    The proof follows from several applications of Rayleigh--Ritz. Throughout these proofs, let $u$ be the top generalized eigenvector of $(H,G)$, and let $v$ be the top eigenvector of $H$. For the first inequality:
    \begin{align*}
    S^{\ell_2} &= \max_{w \neq 0} \frac{\langle w, Hw \rangle}{\langle w, w \rangle}  \\
    &\geq \frac{\langle u, Hu \rangle}{\langle u,u \rangle} \\
    &= S^\Phi \frac{\langle u, Gu \rangle}{ \langle u, u \rangle} \\
    &\geq S^\Phi \minEV{G}.
    \end{align*}
    For the second inequality:
    \begin{align*}
    S^{\Phi} &= \max_{w \neq 0} \frac{\langle w, Hw \rangle}{\langle w, Gw \rangle}  \\
    &\geq \frac{\langle v, Hv \rangle}{\langle v,Gv \rangle} \\
    &= S^{\ell_2} \cdot \frac{\langle v, v \rangle}{ \langle v, Gv \rangle} \\
    &\geq S^{\ell_2} \cdot \frac{1}{\maxEV{G}}.
    \end{align*}

\subsection{Proof of \Cref{prop:sharpness_time_derivative}}

    By chain rule,
    \begin{align*}
        \frac{d}{dt} S^\Phi &= 
        \left\langle \frac{d \bar{z}}{dt} ,  \nabla \widetilde{S}^\Phi(\bar{z})\right\rangle \\
        &= \left\langle -\eta \left( F(\bar{z}) + \frac{\sigma^2}{2} D^2 F(\bar{z})[r,r]\right) ,  \nabla \widetilde{S}^\Phi(\bar{z})\right\rangle \tag{\Cref{eq:DCF}}\\
        &= \eta \langle \nabla \widetilde{S}(\bar{z}), - F(\bar{z}) \rangle - \frac{\eta \sigma^2}{2} \langle \nabla \widetilde{S}(\bar{z}), D^2F(\bar{z})[r,r] \rangle
    \end{align*}

\subsection{Proof of \Cref{lem:negative_beta}}

    The overall idea is that we can arbitrarily pick $H$ and $\nabla^3 L[u,u]$ to be adversarial, where $u$ is the maximum eigenvector that satisfies $SGu = Hu$ and $\langle u, Gu \rangle = 1$, which is equivalent to the problem defined in \Cref{def:sharpness}. Let \[\mathcal{L} = \{\text{losses } L: \|\nabla^3L[\cdot, \cdot, \cdot]\| \text{ bounded}\}.\] 
    
    We present a \emph{affine} reparameterization that allows us to take $G(x^\star)= I$ without loss of generality. Consider the coordinate transformation $x = x^\star + G(x^\star)^{-1/2}y$, and the reparameterized loss and mirror map
    \[\widehat{L}(y) := L\left(x^\star + G(x^\star)^{-1/2}y\right).\]
    \[\widehat{\Phi}(y) := \Phi\left(x^\star + G(x^\star)^{-1/2}y\right).\]
    One may observe that when $x = x^\star$, we have $y = 0$ with
    \[\nabla^2 \widehat{\Phi}(0) = I,\]
    or we find that the mirror metric in the reparameterized coordinates is precisely Euclidean. We can compute that $\beta$ remains unchanged under these coordinates, and $\widehat{L}$ remains in $\mathcal{L}$.

    Without loss of generality, take $G(x^\star) = I$ and $x^\star = 0$. 
    Let $T:=\nabla^3\Phi(x^\star)$. Noting that $T\neq 0$, we claim that there exists a unit vector $u$ such that \[ T[u,u]\notin \operatorname{span}\{u\}. \] 
    To see this, suppose not, and write $f(x) = T[x,x,x]$. Then, one may compute $\nabla f(x) = 3T[x,x]$. We observe that because $\nabla f(x)$ is parallel to $x$ everywhere, $f$ must be constant on a unit sphere. However, because $f$ is odd, the only case must be $f = 0$ everywhere on the sphere. We require $d \geq 2$ for this argument because the unit sphere is connected here. This directly implies $T = 0$, which is a contradiction. Thus, let us decompose $T[u,u] = au + b$ for some $b \perp u$ and $b \neq 0$. 
    
    We construct $H$ by considering its projections on $\text{span}\{u\}$ and $u^\perp$, the orthogonal complement. Fix some $\lambda < S$, and let $L$ be such that the Hessian $H$ satisfies
    \[H = Suu^T + \lambda(I - uu^T).\]
     Now, we may compute
    \begin{align*}
        \stabCoeff^\Phi(x) &= \|\nabla S(x)\|^2_{G^{-1}} + \langle G^{-1} \nabla S, (SI - HG^{-1}) T[u,u] \rangle \\
        &= \|\nabla S(x)\|^2_2 + \langle  \nabla S, (SI - H)(au + b) \rangle 
    \end{align*}
    Observe that the eigenvectors of $H$ are $u$ (with eigenvalue $S$) and the vectors perpendicular to $u$ (with eigenvalue $\lambda$). Thus, $Hu = Su$ and $Hb = \lambda b$, so
    \begin{equation}
        \label{eq:yay}
        \stabCoeff^\Phi(x) = \|\nabla S(x)\|^2_2 + \langle  \nabla S, (S - \lambda) b\rangle.
    \end{equation}
    While keeping the Hessian $H$ fixed at $x^\star$, we may prescribe $\nabla^3L[u,u]$ arbitrarily; for instance, this can be achieved by adding an appropriate cubic polynomial, whose third derivative is constant. By extension, we can control $\nabla S$ arbitrarily by \Cref{lem:sharp_grad}, and we opt to minimize \Cref{eq:yay}.
    We choose $L\in\mathcal L$ so that \[ \nabla S = -\frac{S-\lambda}{2}b. \] Substituting this into the stabilization coefficient gives \begin{align*} \stabCoeff^\Phi &= \frac{(S-\lambda)^2}{4}\|b\|_2^2 - \frac{(S-\lambda)^2}{2}\|b\|_2^2 \\ &= -\frac{(S-\lambda)^2}{4}\|b\|_2^2 <0. \end{align*} 

\subsection{Proof of \Cref{prop:spectral}}

  Observe that $q_i = G^{1/2}u_i$ forms an orthonormal basis in the $L^2$ inner product. Thus, we may decompose:
    \[G^{-1/2}\nabla^3\Phi[u,u] = \sum_i \left\langle G^{-1/2}\nabla^3\Phi[u,u],q_i \right\rangle q_i.\]
    Furthermore, observe that the $q_i$'s are eigenvectors of $G^{-1/2}HG^{-1/2}$ with $G^{-1/2} H G^{-1/2}u_i=\lambda_iu_i$. Expanding $\beta^\Phi(x)$ from \Cref{prop:primal_beta}, we find:
\begin{align*}
    \beta^\Phi(x) &= \|\nabla S\|_{G^{-1}}^2  + \left \langle G^{-1/2} \nabla S, G^{-1/2}(SI- HG^{-1})G^{1/2} G^{-1/2} \nabla^3 \Phi [u,u] \right \rangle \\
    &= \|\nabla S\|_{G^{-1}}^2  + \left\langle G^{-1/2} \nabla S, (SI- G^{-1/2}HG^{-1/2}) G^{-1/2} \nabla^3 \Phi [u,u] \right\rangle \\
        &=
                \|\nabla S\|_{G^{-1}}^2  +\left\langle
            G^{-1/2}\nabla S,
            (SI-G^{-1/2} H G^{-1/2})\sum_i \left\langle G^{-1/2}\nabla^3\Phi[u,u],q_i \right\rangle q_i
        \right\rangle \\
        &=\|\nabla S\|_{G^{-1}}^2  + 
                \sum_i  \left\langle
            G^{-1/2}\nabla S,
            (S-\lambda_i )q_i
        \right\rangle \left\langle G^{-1/2}\nabla^3\Phi[u,u],q_i \right\rangle  \\
        &=\|\nabla S\|_{G^{-1}}^2  + \sum_i
        (S-\lambda_i)
        \langle G^{-1/2}\nabla S,q_i\rangle
        \left\langle
            G^{-1/2}\nabla^3\Phi[u,u],q_i
        \right\rangle\\
        &=\|\nabla S\|_{G^{-1}}^2  + \sum_i
        (S-\lambda_i)
        \langle \nabla S,u_i\rangle
        \left\langle
            \nabla^3\Phi[u,u],u_i
        \right\rangle.
    \end{align*}

\subsection{Auxiliary computations}
\label[appendix]{app:aux_computations}
Throughout this section, we will let $u(x)$ be the top unit eigenvector, and further assume that it is unique. More specifically, $u$ is the maximum eigenvector that satisfies $SGu = Hu$ and $\langle u, Gu \rangle = 1$, as defined in \Cref{def:sharpness}. Define $r = Gu$, which is the pushforward of $u$ under $\nabla \Phi$.

\begin{lemma}[Gradient in dual coordinates]
\label[lemma]{lem:dual_gradient}
Let $f:E\to\mathbb{R}$ and define
\[
\widetilde f := f\circ\nabla\Phi^*.
\]
For $z=\nabla\Phi(x)$,
\[
\nabla \widetilde f(z)
=
\nabla^2\Phi^*(z)\nabla f(x)
=
G(x)^{-1}\nabla f(x).
\]
\end{lemma}

By the chain rule,
\[
\nabla \widetilde f(z)
=
\nabla^2\Phi^*(z)\nabla f(x).
\]
Moreover, differentiating
$\nabla\Phi^*(\nabla\Phi(x))=x$ gives
\[
\nabla^2\Phi^*(z)\nabla^2\Phi(x)=I,
\]
and hence $\nabla^2\Phi^*(z)=G(x)^{-1}$.

\begin{lemma}[Computation of sharpness gradient]
    \label[lemma]{lem:sharp_grad}
    \[\nabla S^{\Phi}(x) = \nabla^3 L(x)[u,u] - S(x) \nabla^3 \Phi(x)[u,u].\]
\end{lemma}

    Define $H:= \nabla^2 L(x)$. Let $u$ satisfy the eigenvalue problem 
    \[SG u = Hu, \qquad \langle u, Gu \rangle = 1\]
    Take gradients on both sides,
    \[(\nabla H) u + H(\nabla u) = (\nabla S)Gu + S(\nabla G) u + SG \nabla u.\]
    Taking an inner product with $u$,
    \begin{align*}
        \langle u, (\nabla H) u\rangle + \langle u, H(\nabla u) \rangle =\nabla S \langle u,Gu  \rangle  + S\langle u,(\nabla G) u  \rangle + S\langle u,G(\nabla u) \rangle
    \end{align*}
    Notice that
    \[\langle u, H(\nabla u)\rangle  = \langle Hu, \nabla u \rangle = S\langle Gu, \nabla u \rangle.\]
    Rearranging and writing the identity in tensor notation yields the result.

\begin{proposition}
    \label[proposition]{prop:curv2curv}
    Let $x \in E, v \in E$, $z = \nabla \Phi (x)$.
    \[\nabla^3 \Phi^* (z) [Gv, Gv] = -G^{-1} \nabla^3\Phi(x)[v,v].\]
\end{proposition}

    It is well-known that the left-hand side can be expressed as the acceleration of a curve \cite{LeeSM}. In particular, let $\alpha(t) = \nabla \Phi^*(z + t Gv)$. Then, $\alpha''(0) = \nabla^3 \Phi^* (z) [Gv, Gv]$.
    By applying $\nabla \Phi$, one can write
    \[\nabla \Phi \circ \alpha(t) = z + t Gv\]
    and taking derivatives
    \begin{align*}
    \nabla^2 \Phi(\alpha(t)) \alpha'(t) &= Gv \\
    \nabla^3\Phi(\alpha(t))[\alpha'(t),\alpha'(t)] +\nabla^2 \Phi(\alpha(t))\alpha''(t) &= 0.
    \end{align*}
    Plugging in $t = 0$ yields $\alpha'(0) = v$ and $\alpha''(0) = -G^{-1} \nabla^3 \Phi[v,v]$.

\begin{proposition}
    \label[proposition]{prop:primal_beta}
    Let $x \in E$ with $z = \nabla \Phi(x)$. Let $\widetilde{S}(z) = S \circ \nabla \Phi^*$. Then,
     \[ \langle \nabla \widetilde{S}(z), D^{2} F[r,r] \rangle = \|\nabla S(x)\|^2_{G^{-1}} + \langle G^{-1} \nabla S, (SI - HG^{-1}) \nabla^3 \Phi[u,u] \rangle\]
     The above quantity is the same as $\stabCoeff^\Phi(x)$ as defined in~\Cref{def:stabCoeff}.
\end{proposition}

    First, we compute $\nabla S(x)$. By \Cref{lem:sharp_grad},
    \[\nabla S = \nabla^3 L[u,u] - S \nabla^3 \Phi[u,u].\]
    Next, differentiating $F = \nabla L \circ \nabla \Phi^*$ twice yields
    \begin{align*}
    D^{2}F[r,r]  &= \nabla^3 L[u,u] - HG^{-1} \nabla^3 \Phi[u,u] \\
    &= \nabla S + S \nabla^3 \Phi [u,u] - HG^{-1} \nabla^3 \Phi[u,u]
    \end{align*}
    By \Cref{lem:dual_gradient}, $\nabla \widetilde{S}(z) = G^{-1} \nabla S(x)$. Putting these together yields:
    \begin{align*}
\langle  \nabla \widetilde{S}(z), D^2F[r,r] \rangle& = \langle G^{-1}\nabla S(x), \nabla S + S \nabla^3 \Phi [u,u] - HG^{-1} \nabla^3 \Phi[u,u]\rangle \\
 &= \|\nabla S(x)\|^2_{G^{-1}} + \langle G^{-1} \nabla S, (SI - HG^{-1}) \nabla^3 \Phi[u,u] \rangle.
\end{align*}

\section{Experimental Details}
\label[appendix]{app:experimetal_details}

Experiments were run on a shared cluster of V100s, A100s, and H200s.
  
% \paragraph{NCI1 \citep{Morris+2020} details} For the GNN experiments, we used a training set consisting of $800$ graphs, which are obtained by taking the first $400$ of the two classes. We use one-hot encodings for node labels and do not use any edge information.

\subsection{Computational details}
\label[appendix]{app:computational_details}

\paragraph{Eigenvalues and sharpness} All eigenvalues are calculated through a LOBPCG algorithm \citep{Kny01} and use the implementation of \cite{CDTKL24} (keeping their condition for termination). This method is particularly nice for computing sharpness because it uses the previous eigenvector to ``warm start'' the solver for the next computation. All computations are done in \texttt{Float64}, as we observe that \texttt{Float32} can incur substantial numerical errors. In our experiments in \Cref{app:more_experiments}, we plot the sharpness every $30$ steps. 

For the $\ell_3$ mirror potential, if any parameter $|x_i| < 10^{-30}$, we set $|x_i| = 10^{-30}$ for the purposes of computing $G^{-1}$ and $G^{-1/2}$, which are singular when $x_i = 0$ as described in \Cref{sec:non-euclidean}. In practice, we observe that this route is never used \emph{except} when layers are initialized at zero. This happens only in our vision transformer, where some parameters are set to zero at initialization.

\paragraph{Central flows} We try to follow the choices of \cite{CDTKL24} in general. For the flows that represent mirror descent, we do all updates in the dual, which we observe to incur less error. The dual central flow is computed using an Euler method, with step size $0.25$ (so that four steps correspond to a mirror descent update, which takes unit time). We discuss our discretization in \Cref{app:discrete} as it is non-trivial; this discretization is used when $S^\Phi > 1.95/\eta$. The other stable flows are implemented with the conservative choice of adaptive step size $1/\max(4, \ceil{2S^{\Phi}})$. We terminate the gradient flow when the step size becomes smaller than $1/1000$.

Some of our choices also differ from \cite{CDTKL24} (other than only dealing with the regime of one unstable eigenvalue). We do not find it necessary to run mirror descent for a few steps before computing the flows, as the mirror sharpness at initialization falls below $2/\eta$. We also compare with the iterates \emph{directly} instead of a ``second-order'' average of the iterates, as it is unclear how to ``average'' the primal iterates in non-Euclidean geometries.

\subsection{Architectural/dataset details}

We use ReLU activation functions for the experiments in \Cref{app:more_experiments}, and we use GeLU activation functions for the central flow experiments to ensure differentiability of the loss.

\paragraph{MLP} We use the MLP architecture of \citet{CKLKT21}. We use PyTorch's standard initializations. This architecture has $655604$ parameters.

\paragraph{CNN} We use the CNN architecture of \citet{CKLKT21}. We use PyTorch's standard initializations. This architecture has $30634$ parameters.

\paragraph{ViT} Our ViT follows the same architecture as that of \citet{CDTKL24}, whose implementation is provided in \citep{lucidrains2024vit}. Following \citet{CDTKL24}, in our experiments, we multiply the initialization weights and biases of the final layer by $0.5$ to reduce the sharpness at initialization. This architecture has $496996$ parameters.

% \paragraph{GCN} We use a two-layer graph convolutional network (GCN) \citep{kipf2017semi} with mean pooling. More specifically, our architectures consists of two convolutional layers of width $50$, a mean pooling layer, a fully-connected layer of $50$, and then a linear layer that outputs scores for the two classes. In total, there are $7102$ parameters.

\paragraph{CIFAR-10 details} We used a training set consisting of $1000$ images, which are obtained by taking the first $250$ of the first four classes of CIFAR-10. We use the standard pre-processing of centering and normalizing the data by subtracting the mean and then dividing by the standard deviation for each channel across the \emph{full} CIFAR-10 dataset.

\section{Varying architectures}
\label[appendix]{app:more_experiments}

In this section, we observe the $2/\eta$ mirror sharpness threshold in a variety of architectures beyond this settings described in the main text. For each setting, we graph the training loss, the gradient norm $\|\nabla L\|_{G^{-1}}^2$, the mirror sharpness $S^\Phi$, and the ratio between $\ell_2$ sharpness and mirror sharpness $S^{\ell_2}/S^\Phi$.

\subsection{MLP}

\subsubsection{MSE}

\begin{figure}[H]
    \centering
    \includegraphics[width=1.0\linewidth]{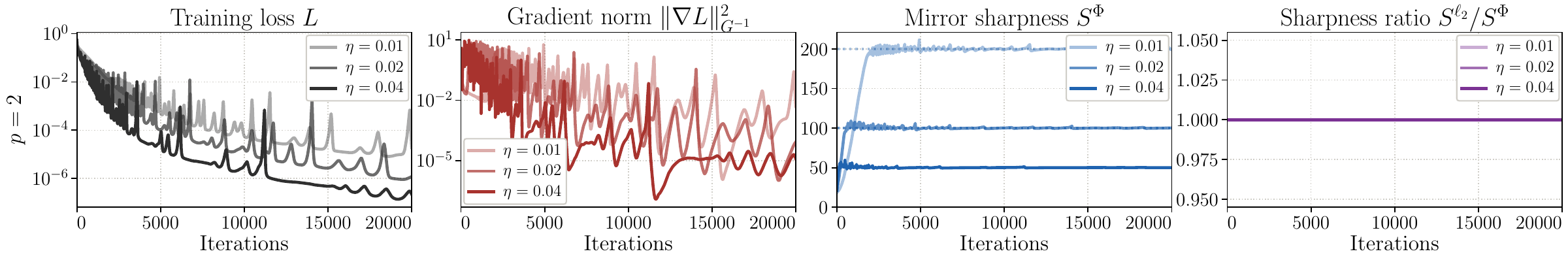}
    \caption{MLP with an MSE loss on a subset of CIFAR-10; ordinary gradient descent.}
    \label{fig:placeholder}
\end{figure}

\begin{figure}[H]
    \centering
    \includegraphics[width=1.0\linewidth]{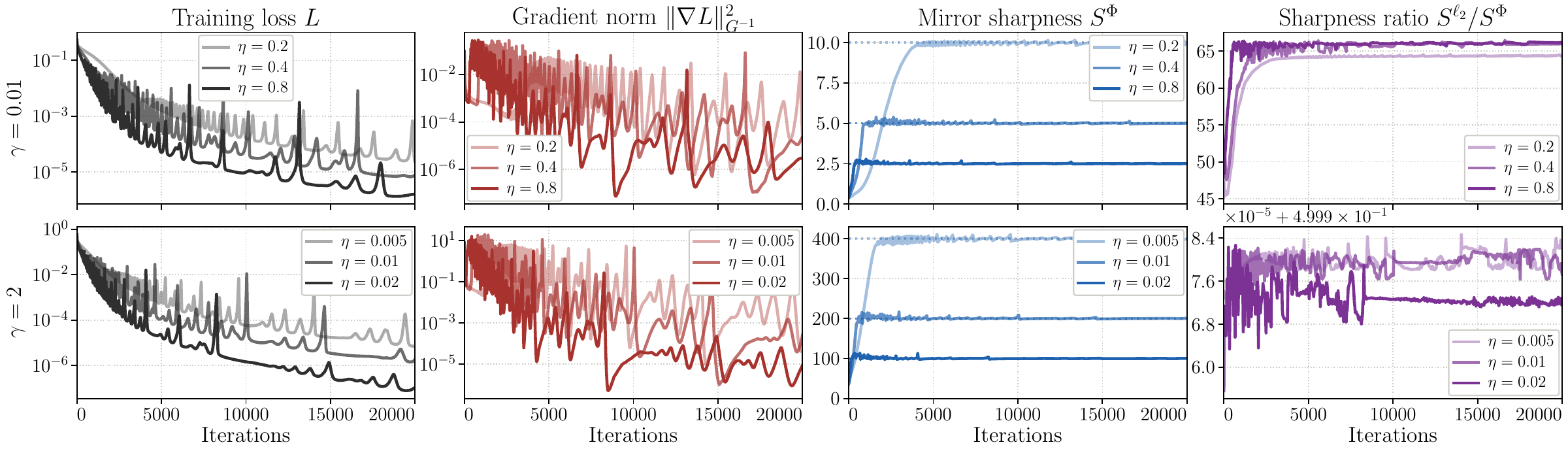}
    \caption{MLP with an MSE loss on a subset of CIFAR-10; hypentropy potential. Top: $\gamma = 0.01$, bottom: $\gamma =2$.}
    \label{fig:placeholder}
\end{figure}

\begin{figure}[H]
    \centering
    \includegraphics[width=1.0\linewidth]{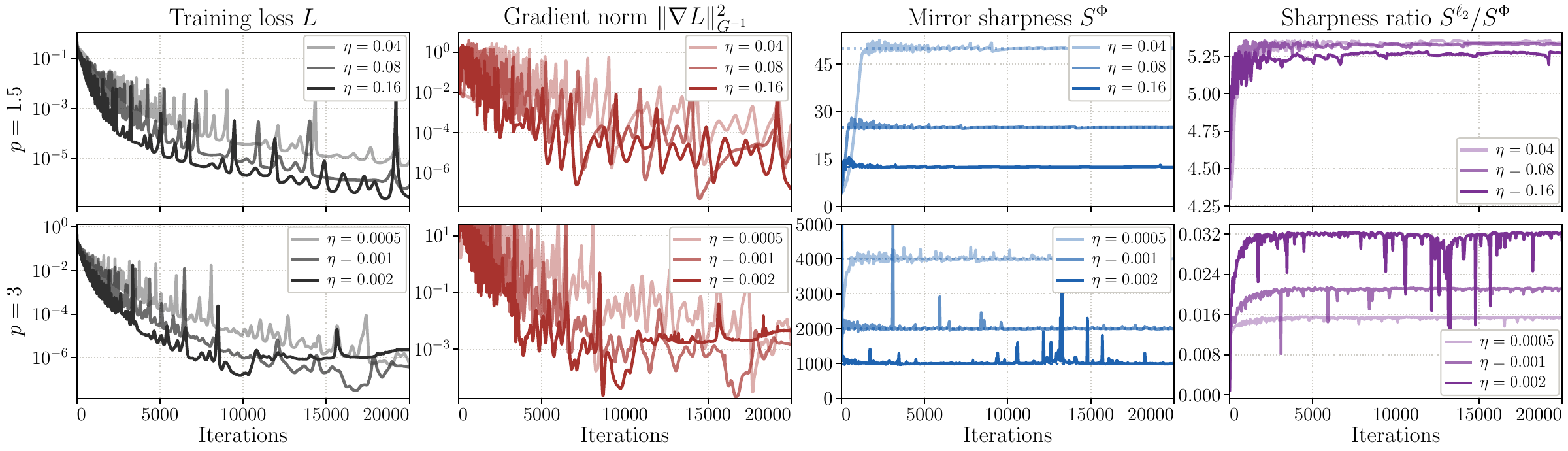}
    \caption{MLP with an MSE loss on a subset of CIFAR-10; $\ell_p$ potential. Top: $\ell_{1.5}$, bottom: $\ell_3$.}
\end{figure}

\subsubsection{Cross-entropy}

For cross-entropy losses, we observe that the mirror sharpness hovers slightly above the $2/\eta$ threshold before returning to zero. This is well-explained in the case for gradient descent in \citet[Appendix C]{CKLKT21}, and we anticipate that a similar explanation may be at play here.

\begin{figure}[H]
    \centering
    \includegraphics[width=1.0\linewidth]{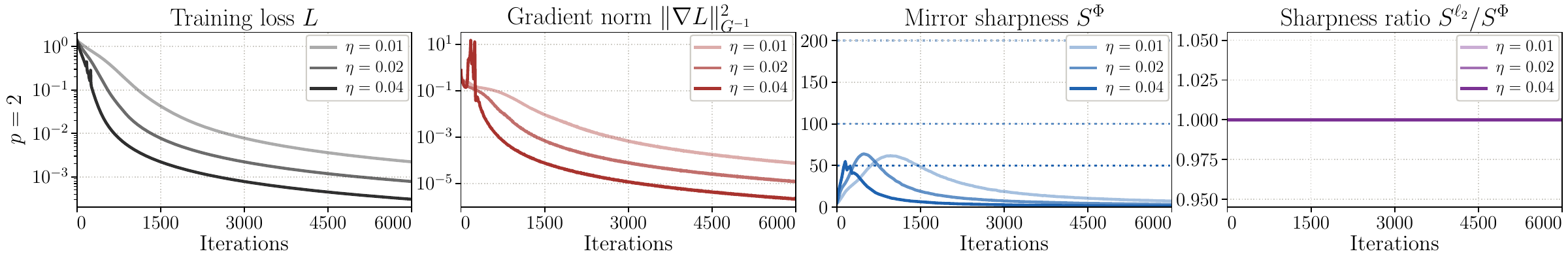}
    \caption{MLP with a cross-entropy loss on a subset of CIFAR-10; ordinary gradient descent.}
    \label{fig:placeholder}
\end{figure}

\begin{figure}[H]
    \centering
    \includegraphics[width=1.0\linewidth]{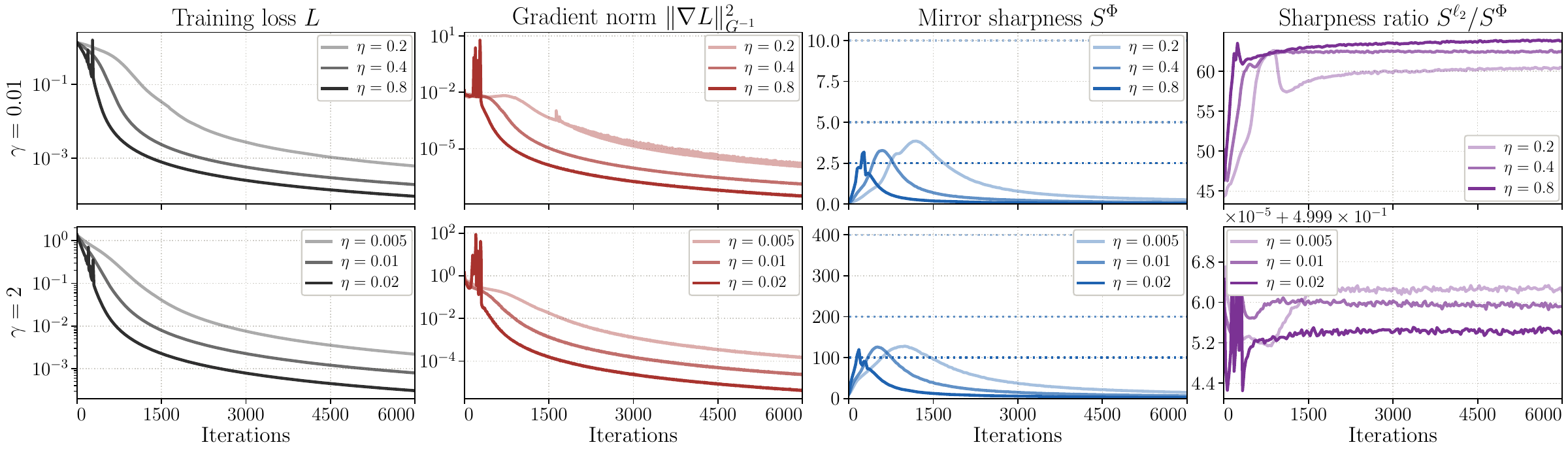}
    \caption{MLP with a cross-entropy loss on a subset of CIFAR-10; hypentropy potential. Top: $\gamma = 0.01$, bottom: $\gamma =2$.}
    \label{fig:placeholder}
\end{figure}

\begin{figure}[H]
    \centering
    \includegraphics[width=1.0\linewidth]{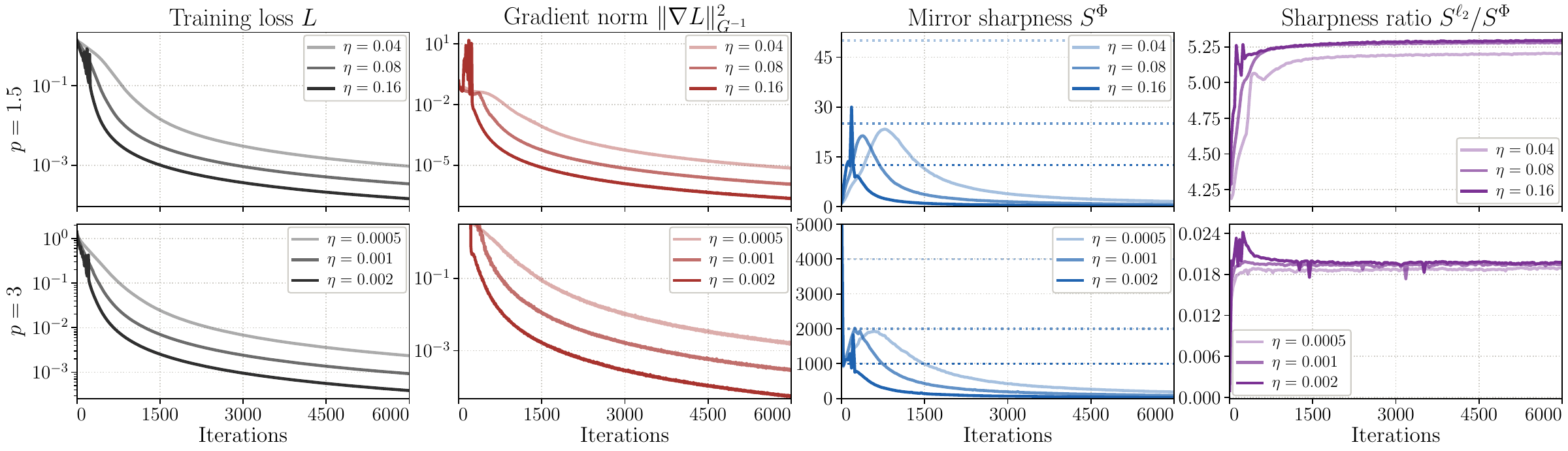}
    \caption{MLP with a cross-entropy loss on a subset of CIFAR-10; $\ell_p$ potential. Top: $\ell_{1.5}$, bottom: $\ell_3$.}
\end{figure}

\subsection{CNN}

\subsubsection{MSE}

\begin{figure}[H]
    \centering
    \includegraphics[width=1.0\linewidth]{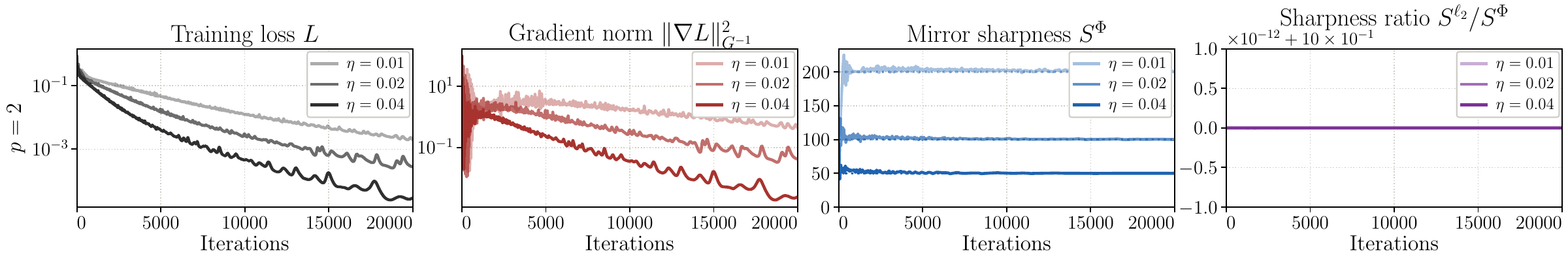}
    \caption{CNN with an MSE loss on a subset of CIFAR-10; gradient descent.}
    \label{fig:CNN_pnorm_CE}
\end{figure}

\begin{figure}[H]
    \centering
    \includegraphics[width=1.0\linewidth]{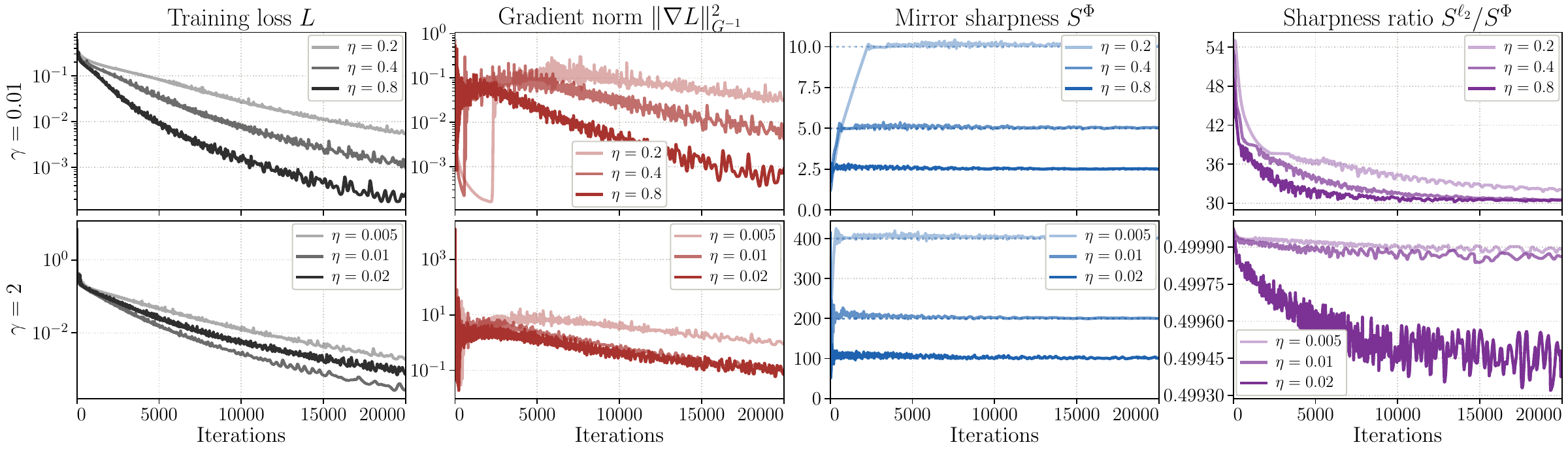}
    \caption{CNN with an MSE loss on a subset of CIFAR-10; hypentropy potential. Top: $\gamma = 0.01$, bottom: $\gamma =2$. These are the same experiments as \Cref{fig:main_text_cnn_hyp_mse}.}
    \label{fig:placeholder}
\end{figure}

\begin{figure}[H]
    \centering
    \includegraphics[width=1.0\linewidth]{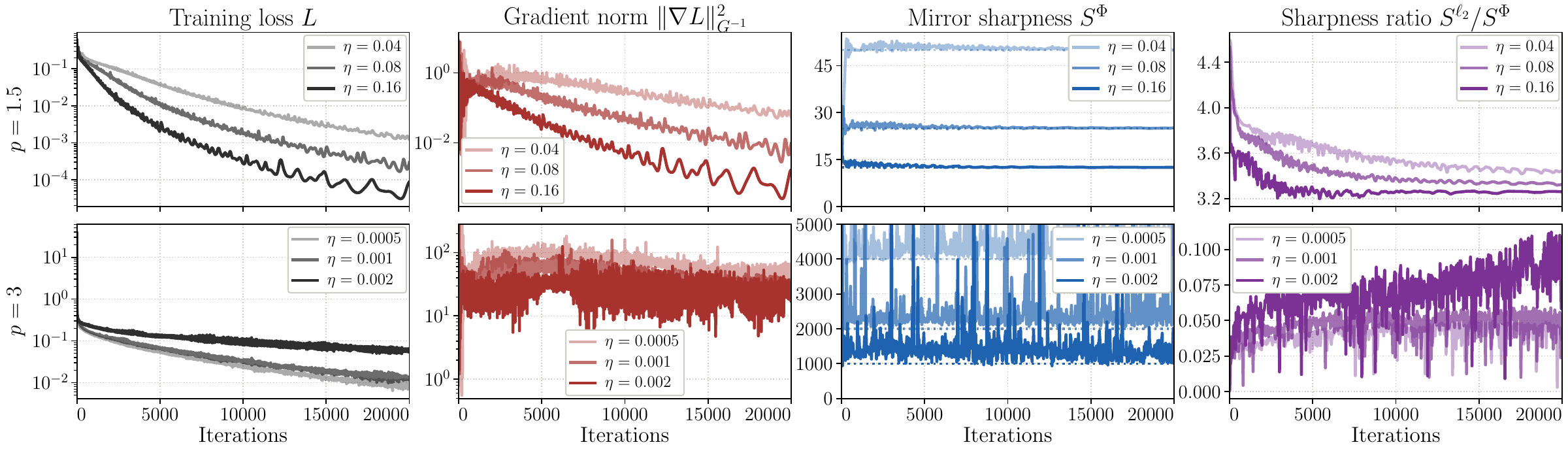}
    \caption{CNN with an MSE loss on a subset of CIFAR-10; $\ell_p$ potential. Top: $\ell_{1.5}$, bottom: $\ell_3$.}
    \label{fig:CNN_pnorm_CE}
\end{figure}

\subsubsection{Cross-entropy}

\begin{figure}[H]
    \centering
    \includegraphics[width=1.0\linewidth]{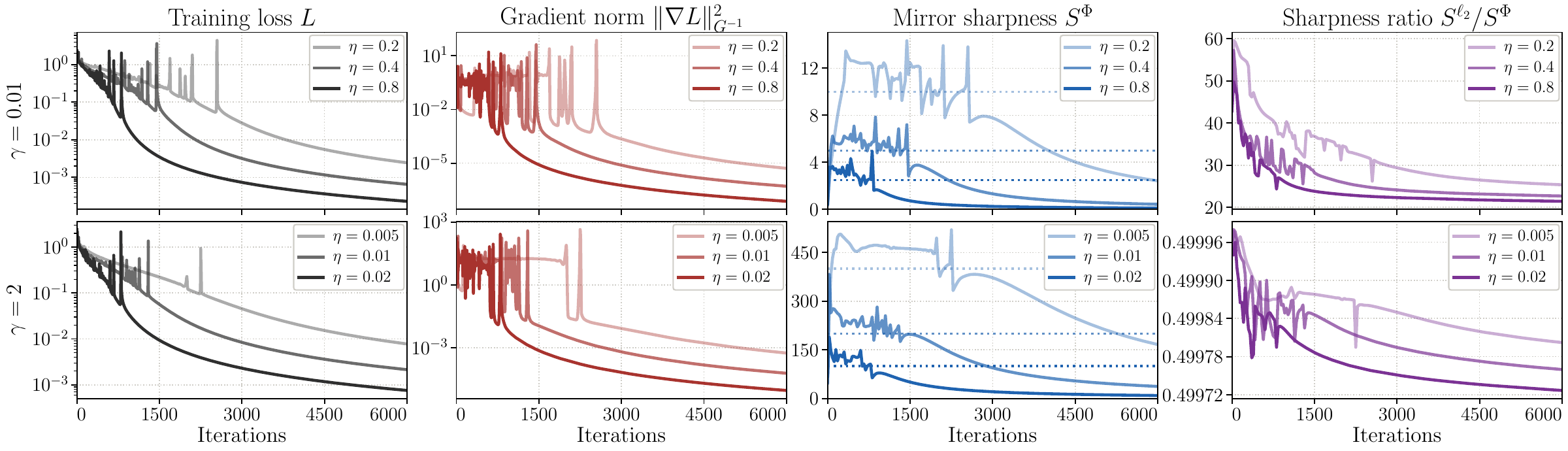}
    \caption{CNN with cross-entropy loss on a subset of CIFAR-10; hypentropy potential. Top: $\gamma = 0.01$, bottom: $\gamma =2$.}
    \label{fig:CNN_hyp_CE}
\end{figure}

\begin{figure}[H]
    \centering
    \includegraphics[width=1.0\linewidth]{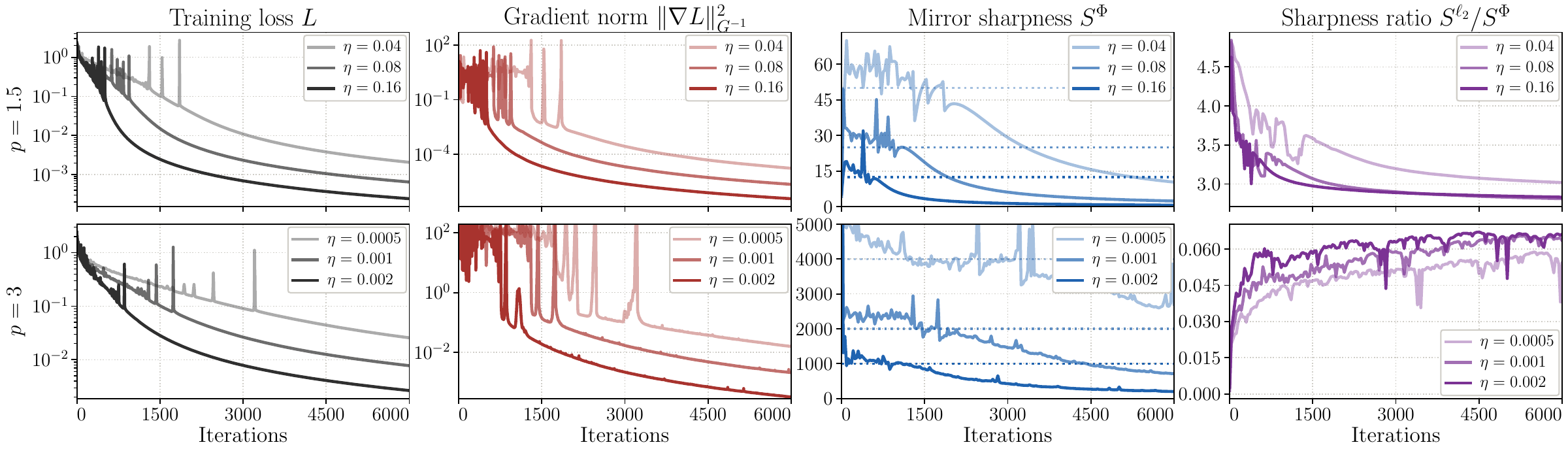}
    \caption{CNN with cross-entropy loss on a subset of CIFAR-10; $\ell_p$ potential. Top: $\ell_{1.5}$, bottom: $\ell_3$.}
    \label{fig:CNN_pnorm_CE}
\end{figure}

\subsection{ViT}
\subsubsection{MSE}
\begin{figure}[H]
    \centering
    \includegraphics[width=1.0\linewidth]{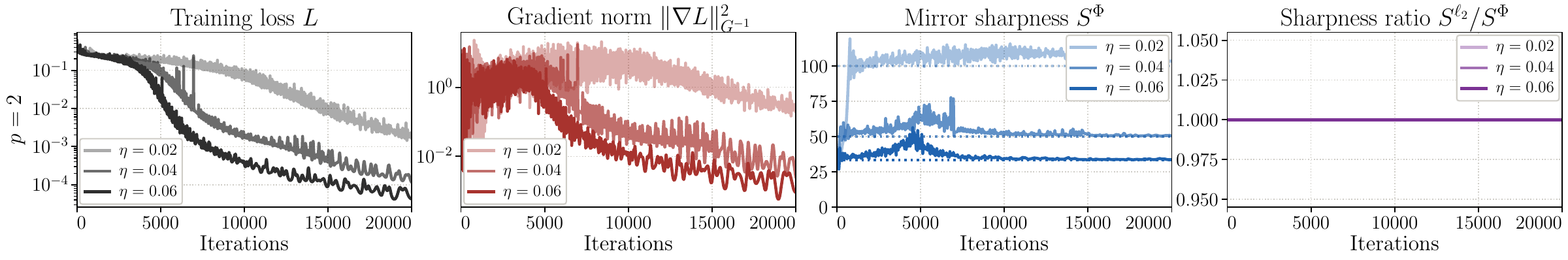}
    \caption{ViT with an MSE loss on a subset of CIFAR-10; ordinary gradient descent.}
    \label{fig:placeholder}
\end{figure}

\begin{figure}[H]
    \centering
    \includegraphics[width=1.0\linewidth]{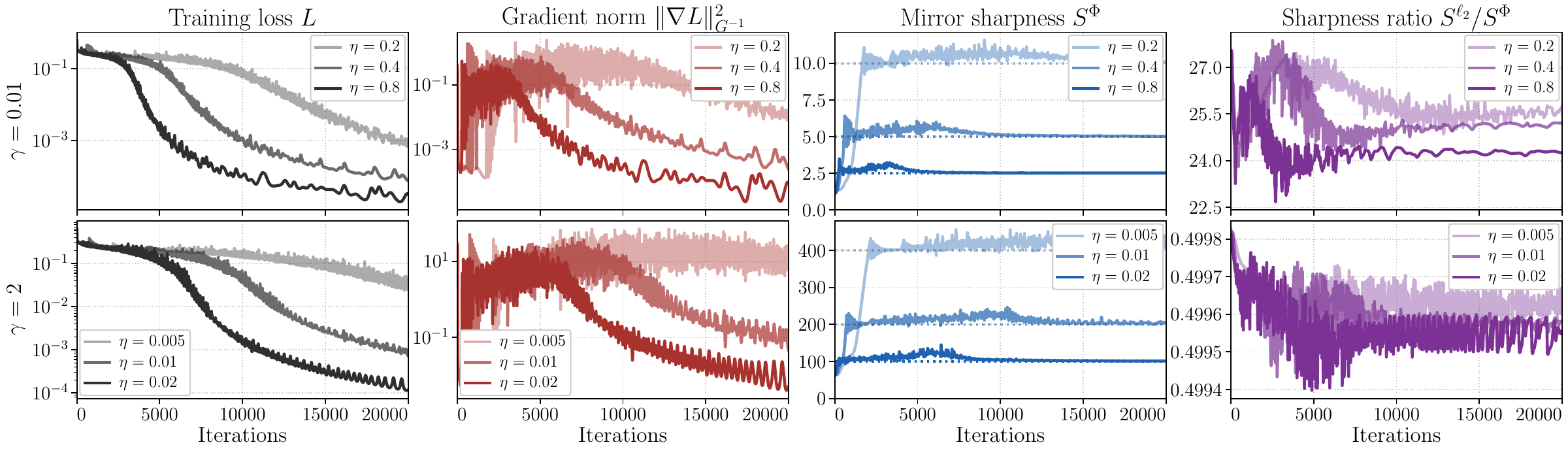}
    \caption{ViT with an MSE loss on a subset of CIFAR-10; hypentropy potential. Top: $\gamma = 0.01$, bottom: $\gamma =2$.}
\end{figure}

\begin{figure}[H]
    \centering
    \includegraphics[width=1.0\linewidth]{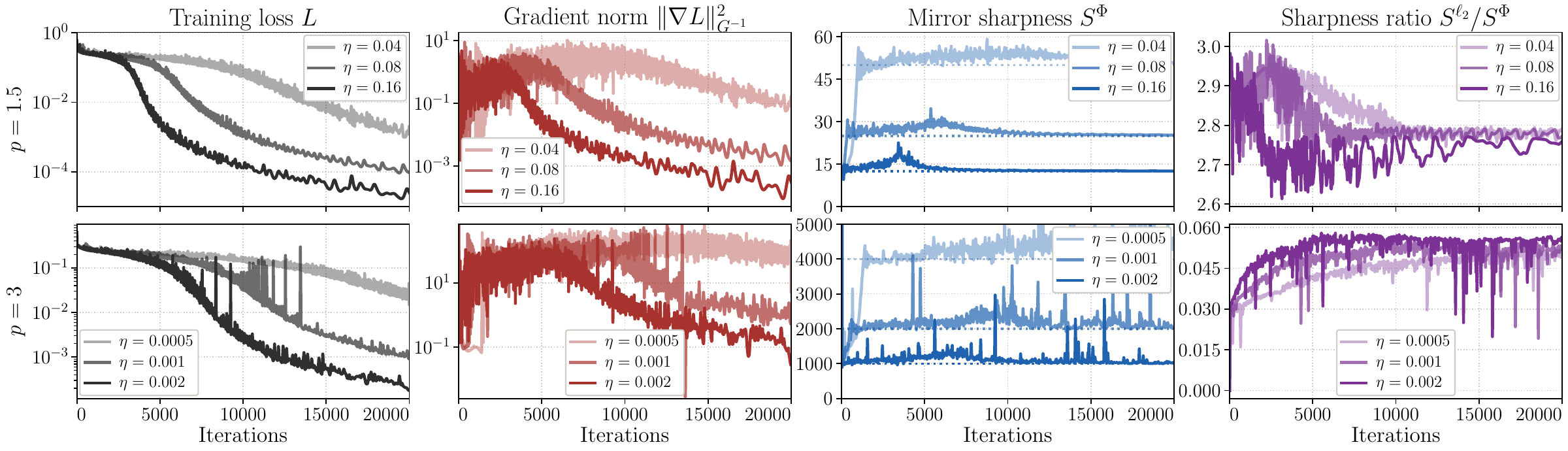}
    \caption{ViT with an MSE loss on a subset of CIFAR-10; $\ell_p$ potential. Top: $\ell_{1.5}$, bottom: $\ell_3$. These are the same experiments as \Cref{fig:main_text_vit_pnorm_mse}.}
\end{figure}

\subsubsection{Cross-entropy}

\begin{figure}[H]
    \centering
    \includegraphics[width=1.0\linewidth]{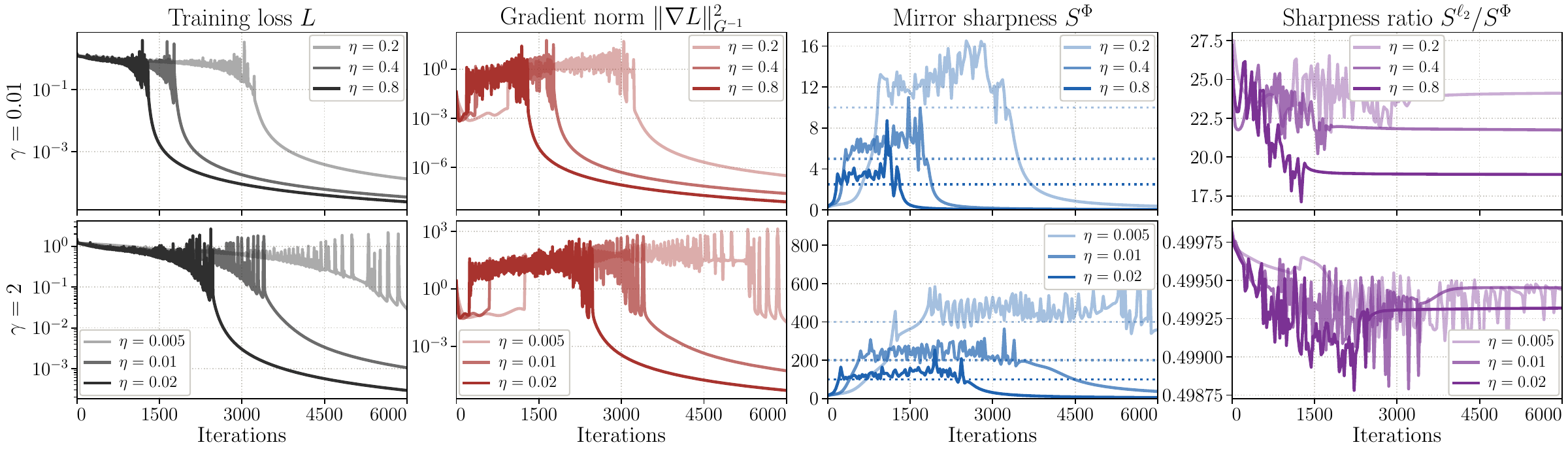}
    \caption{ViT with cross-entropy loss on a subset of CIFAR-10; hypentropy potential. Top: $\gamma = 0.01$, bottom: $\gamma =2$.}
    \label{fig:vit_hyp_CE}
\end{figure}

\begin{figure}[H]
    \centering
    \includegraphics[width=1.0\linewidth]{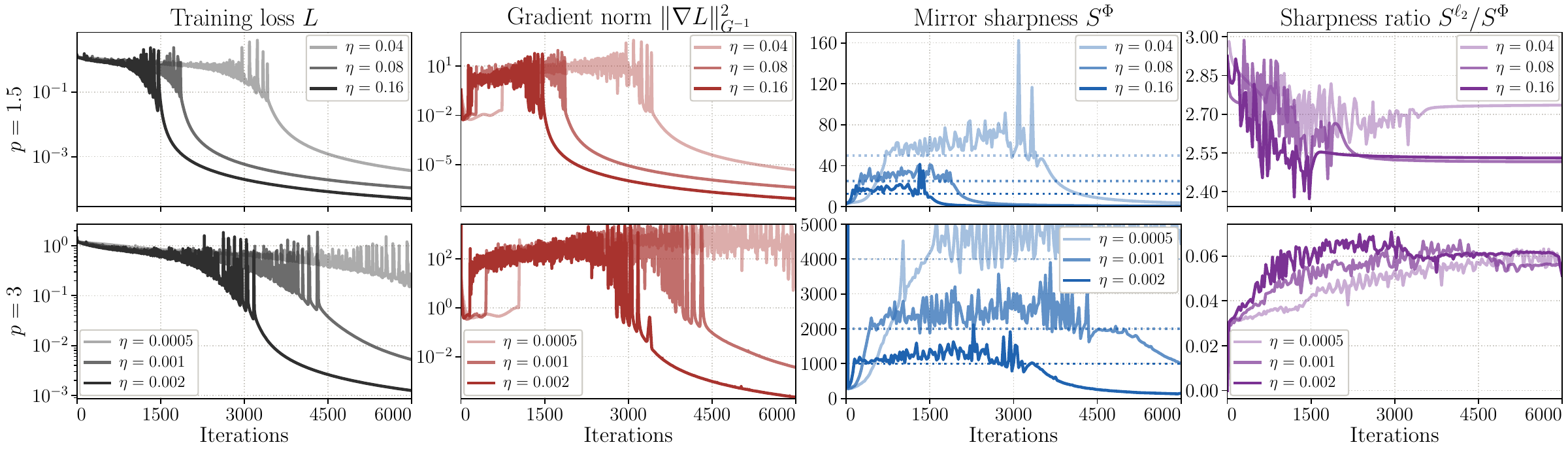}
    \caption{ViT with cross-entropy loss on a subset of CIFAR-10; $\ell_p$ potential. Top: $\ell_{1.5}$, bottom: $\ell_3$.}
    \label{fig:vit_pnorm_CE}
\end{figure}

\section{Central flow experiments}
\label[appendix]{app:ce_experiments}

\begin{figure}[H]
    \centering
    \includegraphics[width=1.0\linewidth]{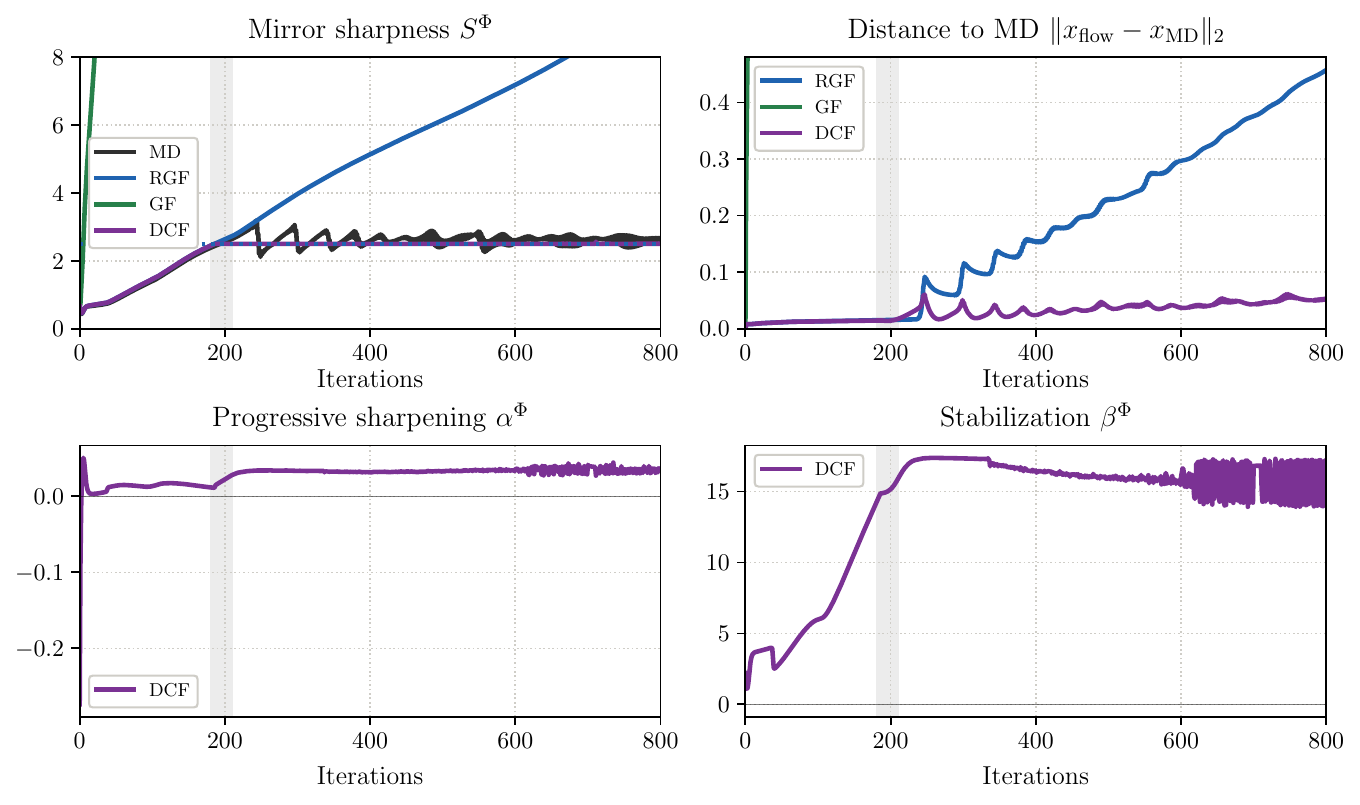}
    \caption{CNN with an MSE loss on a subset of CIFAR-10; hypentropy potential with $\gamma = 0.01$; $\eta = 0.8$. We indicate, in gray, the timesteps where the DCF iterates only have one unstable eigenvector with eigenvalue greater than $1.95/\eta$.}
    \label{fig:CNN_hypentropy_MSE_CF}
\end{figure}

\begin{figure}[H]
    \centering
    \includegraphics[width=1.0\linewidth]{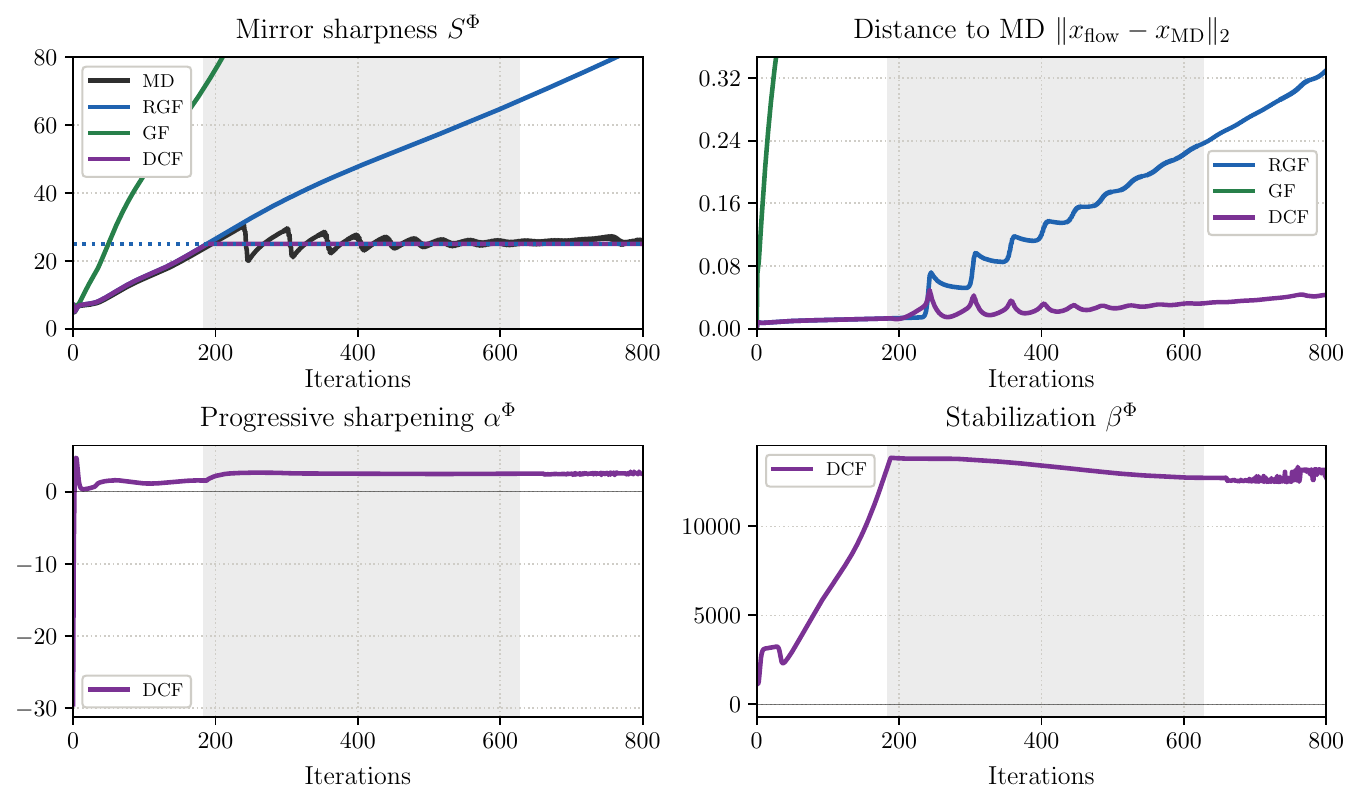}
    \caption{CNN with an MSE loss on a subset of CIFAR-10; $\ell_{1.5}$ potential; $\eta = 0.08$. This is the same experiment as \Cref{fig:main_text_central_flows}. We indicate, in gray, the timesteps where the DCF iterates only have one unstable eigenvector with eigenvalue greater than $1.95/\eta$.}
    \label{fig:CNN_l1.5_MSE_CF}
\end{figure}

\begin{figure}[H]
    \centering
    \includegraphics[width=1.0\linewidth]{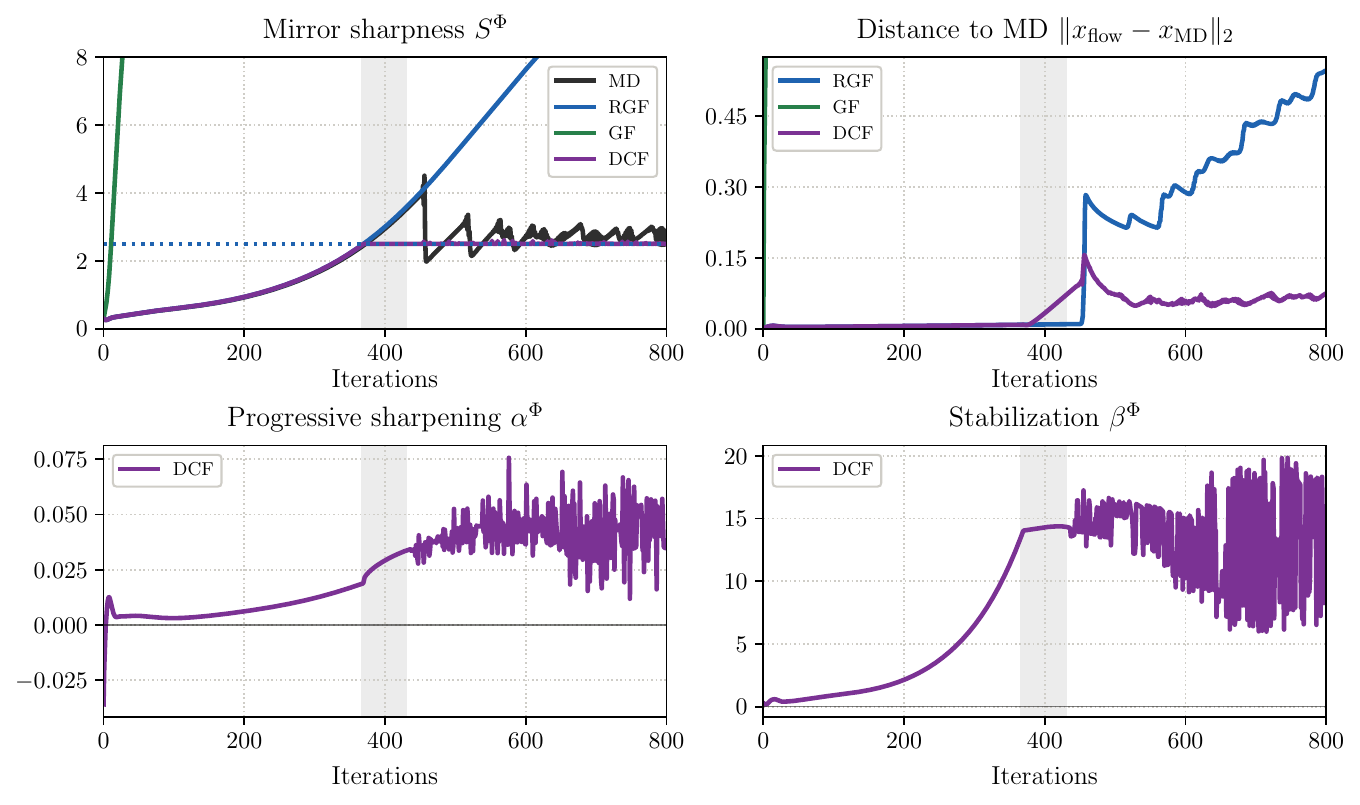}
    \caption{MLP with an MSE loss on a subset of CIFAR-10; hypentropy potential with $\gamma = 0.01$; $\eta = 0.8$. We indicate, in gray, the timesteps where the DCF iterates only have one unstable eigenvector with eigenvalue greater than $1.95/\eta$.}
    \label{fig:MKO_hypentropy_MSE_CF}
\end{figure}

\begin{figure}[H]
    \centering
    \includegraphics[width=1.0\linewidth]{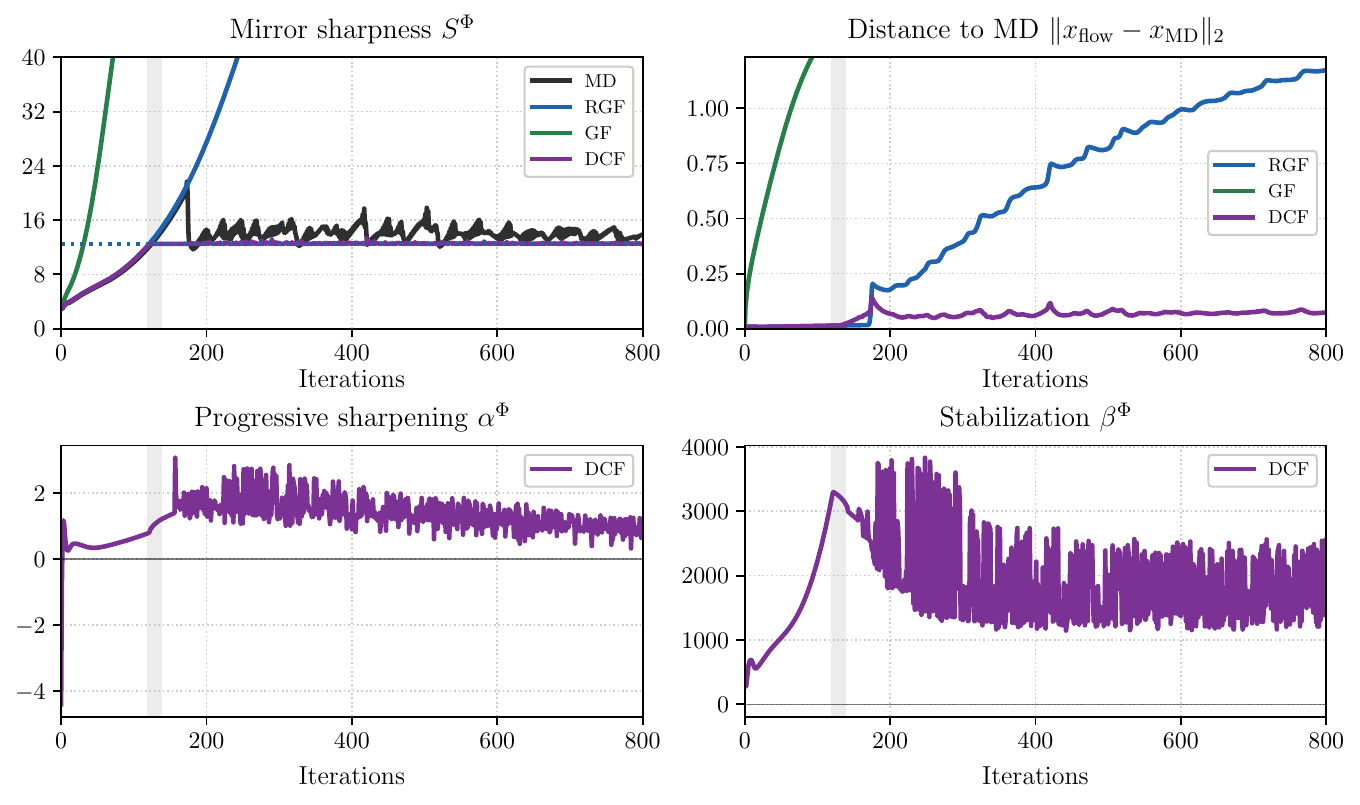}
    \caption{MLP with an MSE loss on a subset of CIFAR-10; $\ell_{1.5}$ potential; $\eta = 0.08$. We indicate, in gray, the timesteps where the DCF iterates only have one unstable eigenvector with eigenvalue greater than $1.95/\eta$.}
    \label{fig:MLP_pnorm_MSE_CF}
\end{figure}

\section{Transformer on Tiny Shakespeare}

So far, we have considered an image classification task with several architectures, mirror maps, losses, and step sizes. In this section, we instead consider a language modeling task. More specifically, we train a transformer to predict the next character in the Tiny Shakespeare dataset \citep{karpathy2015charrnn}. We defer experimental details to \Cref{app:shakespeare_details}.

\begin{figure}[H]
    \centering
    \includegraphics[width=1.0\linewidth]{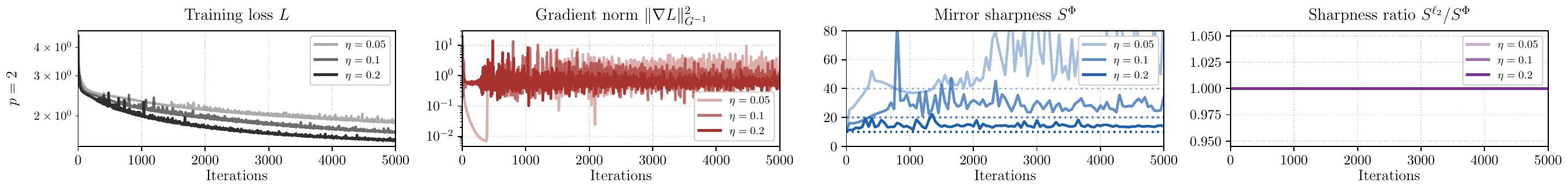}
    \caption{Transformer with a cross-entropy loss on Tiny Shakespeare; gradient descent.}
    \label{fig:CNN_pnorm_CE}
\end{figure}

\begin{figure}[H]
    \centering
    \includegraphics[width=1.0\linewidth]{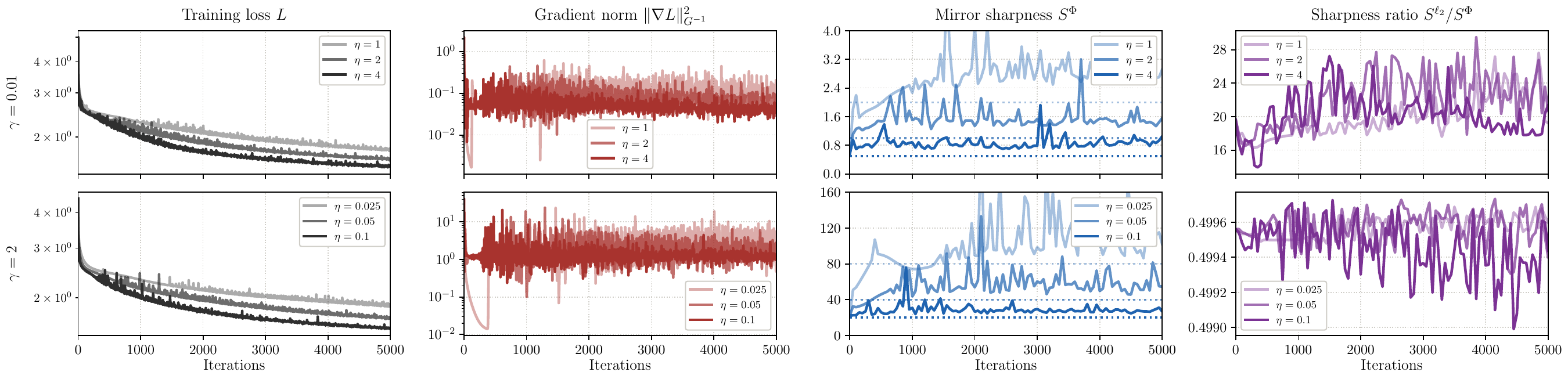}
    \caption{Transformer with a cross-entropy loss on Tiny Shakespeare; hypentropy potential. Top: $\gamma = 0.01$, bottom: $\gamma =2$.}
    \label{fig:CNN_hyp_CE}
\end{figure}

\begin{figure}[H]
    \centering
    \includegraphics[width=1.0\linewidth]{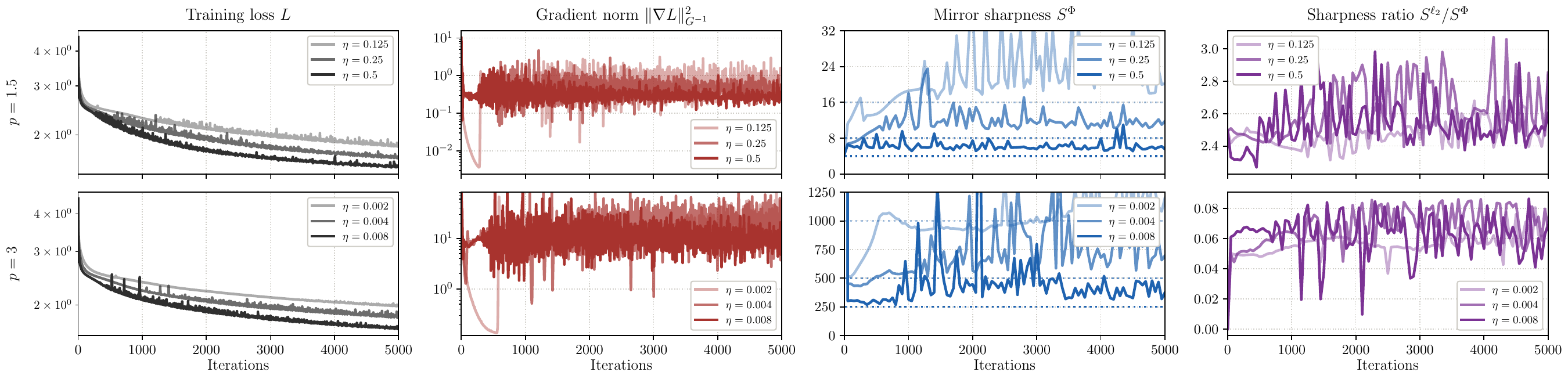}
    \caption{Transformer with a cross-entropy loss on Tiny Shakespeare; $\ell_p$ potential. Top: $\ell_{1.5}$, bottom: $\ell_3$.}
    \label{fig:CNN_pnorm_CE}
\end{figure}

\subsection{Experimental details}
\label[appendix]{app:shakespeare_details}

Our experiment largely follows that of \citet{CKLKT21} (Appendix~L.1), with the Tiny Shakespeare dataset instead of WikiText-2. We use the first 90\% of the text for the training split. We cut this text into sequences of 35 characters, which consist of our training samples. For each sample, the model is asked to predict the next character given the previous. We use a cross-entropy loss, and our architecture is identical to that of \citet{CKLKT21}.

\section{Concrete example of \Cref{lem:negative_beta}}
\label[appendix]{app:numerical_example}

We present a concrete numerical example of \Cref{lem:negative_beta} over parameters in $\mathbb{R}^2$; the main idea is to write out the quadratic and cubic terms separately about a point $x^\star$ and prescribe our desired quantities explicitly.

Consider the hypentropy mirror map with $\gamma = 0.6$, and let $x^\star = (0.8, -0.8)$. This forces $G = \nabla^2\Phi(x^\star) = I$. 

Pick vectors $u = (1/\sqrt{2}, 1/\sqrt{2})$ and $v = (-1/\sqrt{2}, 1/\sqrt{2})$. One may explicitly compute $\nabla^3\Phi(x^\star)[u,u] = (-0.4, 0.4)$, noting that $v$ and $\nabla^3\Phi(x^\star)[u,u]$ point in the same direction.

Fix some $S > 0$. We choose the loss so that the third derivative points along this direction (orthogonal to $u$): for $d \in \mathbb{R}^2$, define
\[L(x^\star + d) = \frac{1}{2} \langle d, S u u^T d \rangle + \frac{S}{4} \langle u, d \rangle^2  \langle \nabla^3 \Phi(x^\star)[u,u], d\rangle,\] 
which satisfies the conditions needed in the proof of \Cref{lem:negative_beta}, as one may check
\[\nabla^2 L(x^\star) =Suu^T \]
\[\nabla^3L(x^\star)[u,u] = \frac{S}{2} \nabla^3\Phi(x^\star)[u,u],\]
noting that we have set $\lambda = 0$ to simplify. Following the computation in the proof of \Cref{lem:negative_beta}, one can check that $\beta^\Phi(x^\star) = -0.08S^2 < 0$.

Some observations about $L$: first, note that $L$ is unbounded below and thus can tend to $-\infty$ in our example. Indeed, in our optimization example, we see that the trajectory runs out of this neighborhood (and in fact the iterates become infinite on our machine). We believe that it is possible to repair this by considering a quartic polynomial with the same information at the first three orders. Second: we note that $x^\star$ is a local minima, with $L(x^\star) = 0$. Thus, we must slightly perturb it, as it would otherwise become a fixed point. Third: we note that we must take $S > 2/\eta$ so that our trajectory moves \emph{away} from $x^\star$.

We consider mirror descent with parameters $S = 16$, $\eta = 0.13$, $\gamma = 0.6$, starting at a point $x_0 = x^\star + 0.01 u$. On our machine, the iterate became infinite after $48$ steps.

\begin{figure}[H]
    \centering
    \includegraphics[width=0.6\linewidth]{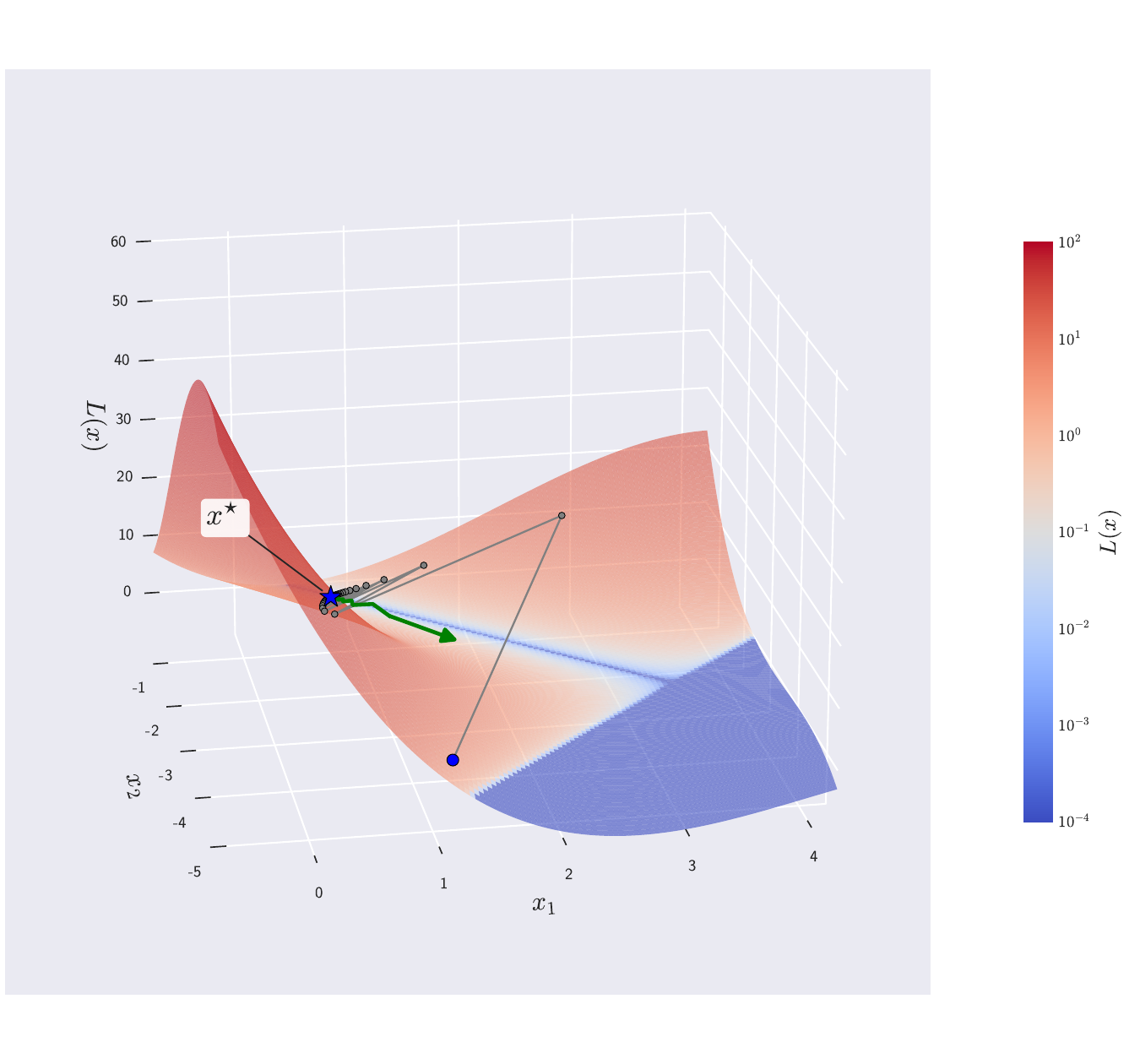}
    \caption{Three-dimensional visualization of the mirror descent iterates on the loss surface $L$ before the iterates diverge. We plot the motion of the average (dual) iterate over the last $6$ steps in green.}
    \label{fig:CNN_pnorm_CE}
\end{figure}

As the central flow model ceases to be relevant since $\beta^\Phi < 0$, we instead track the evolution of the time-averaged iterates by averaging the last six iterates in the dual space. We track $\alpha^\Phi$ and $\beta^\Phi$ for these iterates. We find that for most of the trajectory, $\alpha^\Phi$ remains positive and $\beta^\Phi$ remains negative.

\begin{figure}[H]
    \centering
    \includegraphics[width=1\linewidth]{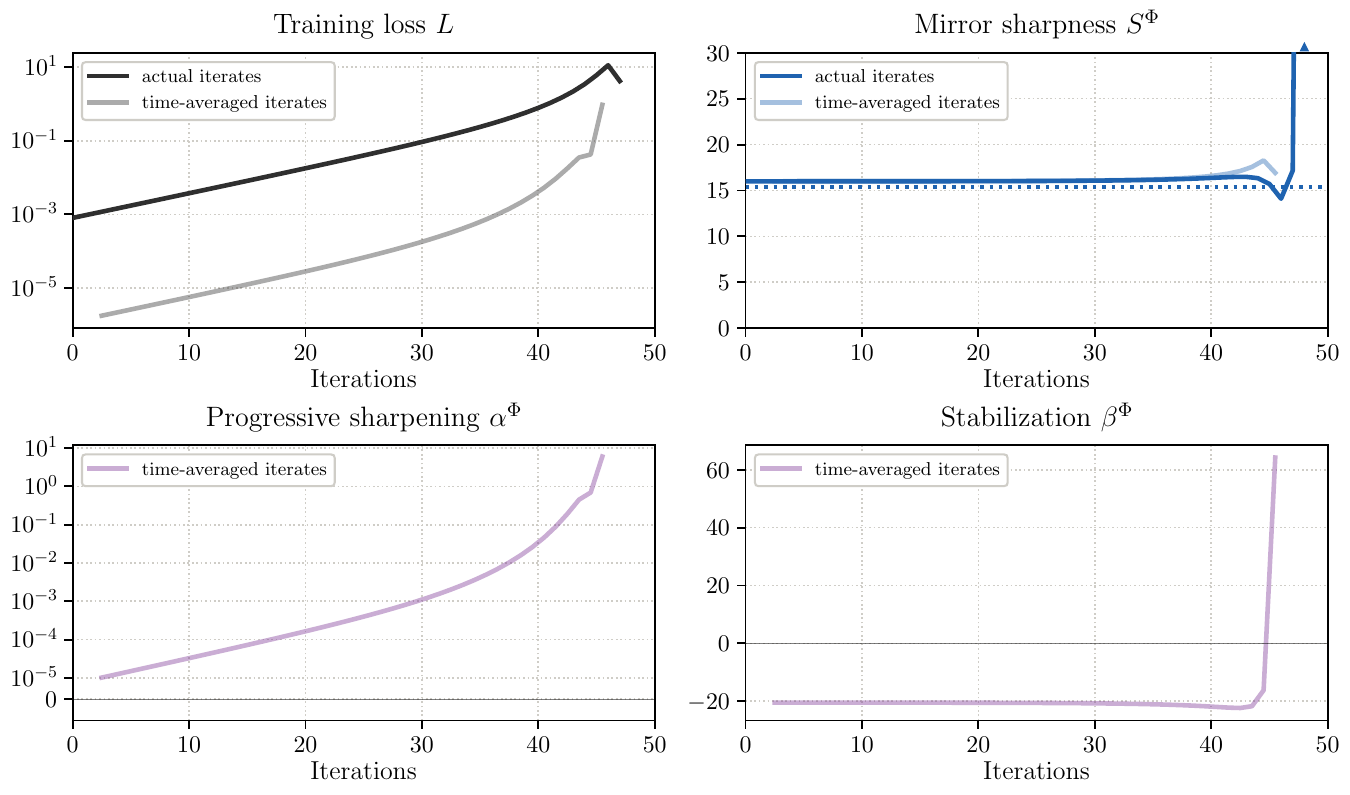}
    \caption{We plot associated quantities of the actual iterates and the time-averaged iterates before divergence.}
    \label{fig:CNN_pnorm_CE}
\end{figure}

\end{document}